%% file: main.tex
\documentclass[11pt,letterpaper]{article}
\PassOptionsToPackage{hyphens}{url}
\PassOptionsToPackage{dvipsnames}{xcolor}
\usepackage[T1]{fontenc}
\usepackage{lmodern}
\usepackage[margin=1in]{geometry}
\usepackage{microtype}
\usepackage{etoolbox}
\usepackage[font=small,labelfont=bf]{caption}
\usepackage{amsmath,amssymb,amsthm,mathtools}
\usepackage{booktabs,tabularx,array,multirow}
\newcolumntype{P}[1]{>{\raggedright\arraybackslash}p{#1}}
\newcolumntype{C}[1]{>{\centering\arraybackslash}p{#1}}
\newcolumntype{L}{>{\raggedright\arraybackslash}X}

\usepackage{graphicx}
\usepackage{xcolor}
\usepackage{placeins}
\usepackage{upquote}
\usepackage{enumitem}
\setlist{itemsep=2pt,topsep=4pt}
\usepackage{tikz}
\usetikzlibrary{positioning,arrows.meta,calc,fit,backgrounds,shapes.geometric,shapes.misc,decorations.pathreplacing}
\usepackage{pgfplots}
\tikzset{every picture/.append style={execute at begin picture={\selectfont}},
  every node/.append style={execute at begin node={\setlength{\baselineskip}{1.18em}}}}
\pgfplotsset{compat=1.17}
\usepackage{listings}
\lstdefinestyle{paperpython}{
 language=Python,
 basicstyle=\ttfamily\fontsize{9}{11.5}\selectfont,
 keywordstyle=\bfseries,
 commentstyle=\itshape\color{black!60},
 columns=fullflexible,keepspaces=true,showstringspaces=false,
 upquote=true,breaklines=false,
 frame=single,framesep=6pt,rulecolor=\color{black!35},
 backgroundcolor=\color{black!2},
 xleftmargin=7pt,xrightmargin=7pt,
 aboveskip=9pt,belowskip=9pt,captionpos=t
}
\lstdefinestyle{memtext}{
 basicstyle=\ttfamily\fontsize{7.6}{9.2}\selectfont,
 columns=fullflexible,keepspaces=true,showstringspaces=false,
 upquote=true,breaklines=true,breakatwhitespace=true,
 frame=single,framesep=4pt,rulecolor=\color{black!30},
 xleftmargin=5pt,xrightmargin=5pt,aboveskip=6pt,belowskip=6pt,captionpos=t
}
\newcommand{\paperbox}[2]{%
\par\vspace{6pt}\begingroup
\setlength{\fboxsep}{8pt}\setlength{\fboxrule}{0.4pt}
\noindent\fcolorbox{black!35}{black!2}{\begin{minipage}{\dimexpr\linewidth-2\fboxsep-2\fboxrule\relax}
\fontsize{10}{13.5}\selectfont
\setlength{\parindent}{0pt}\setlength{\parskip}{4pt}
\setlength{\abovedisplayskip}{5pt}\setlength{\belowdisplayskip}{5pt}
\textbf{#1}\par #2\end{minipage}}
\endgroup\par\vspace{8pt}}
\usepackage{natbib}
\bibpunct[, ]{(}{)}{,}{a}{}{,}

\usepackage{xurl}
\usepackage[hidelinks]{hyperref}

\definecolor{rteal}{RGB}{31,111,101}    
\definecolor{rblue}{RGB}{47,91,140}     
\definecolor{rgray}{RGB}{88,96,108}     
\definecolor{rred}{RGB}{160,52,52}      
\definecolor{rgreen}{RGB}{52,112,64}    

\newcommand{\Pref}{P_{\mathrm{ref}}}
\newcommand{\Pdisp}{P_{\mathrm{disp}}}

\newcommand{\cR}{\mathcal{R}}
\newcommand{\cX}{\mathcal{X}}
\newcommand{\cG}{\mathcal{G}}
\newcommand{\cU}{\mathcal{U}}
\newcommand{\cP}{\mathcal{P}}

\newcommand{\val}{\operatorname{val}}
\newcommand{\OA}{\mathrm{OA}}
\newcommand{\EP}{\mathrm{EP}}
\newcommand{\Dt}{\mbox{D2$_{\theta}$}}
\newcommand{\Fbar}{\bar F}
\newcommand{\cD}{\mathcal{D}}

\newcommand{\ThetaPi}{\Theta_{\Pi}}
\newcommand{\E}{\mathbb{E}}
\newcommand{\Prob}{\mathbb{P}}
\newcommand{\R}{\mathbb{R}}

\newcommand{\parhead}[1]{\smallskip\noindent\textbf{#1}\enspace}

\theoremstyle{plain}
\newtheorem{proposition}{Proposition}
\theoremstyle{definition}
\newtheorem{example}{Example}
\newtheorem{definition}{Definition}
\newtheoremstyle{restatement}{6pt}{6pt}{\itshape}{}{\bfseries}{}{0.5em}{\thmnote{#3}}
\theoremstyle{restatement}
\newtheorem*{repeatproposition}{}
\let\standardproof\proof
\renewcommand{\proof}[1]{\standardproof[#1]}
\newcommand{\Halmos}{\qedhere}

\AtBeginEnvironment{table}{\small\setlength{\tabcolsep}{4pt}}
\hypersetup{
  pdftitle={RobustOPT: A Human--AI Collaborative Modeling System for Robust and Distributionally Robust Optimization},
  pdfauthor={Shuang Jin, Jinbiao Chen, Guoyun Zhang, Yue Zhao, Junyu Zhang, Guanyi Wang, Hanzhang Qin},
  pdfsubject={Robust and distributionally robust optimization},
  pdfkeywords={robust optimization, distributionally robust optimization, human-AI collaborative modeling, reformulation verification}
}

\title{\fontsize{16}{20}\selectfont\bfseries RobustOPT: A Human--AI Collaborative Modeling System\\for Robust and Distributionally Robust Optimization}
\author{%
  \normalsize Shuang Jin$^{1}$\quad Jinbiao Chen$^{1}$\quad Guoyun Zhang$^{2,3}$\\[3pt]
  \normalsize Yue Zhao$^{4}$\quad Junyu Zhang$^{1}$\quad Guanyi Wang$^{1,*}$\quad Hanzhang Qin$^{1,2,*}$\\[9pt]
  \parbox{\textwidth}{\centering\small
    $^1$Department of Industrial Systems Engineering and Management,\\
    National University of Singapore\\[2pt]
    $^2$Institute of Operations Research and Analytics, National University of Singapore\\[2pt]
    $^3$Agency for Science, Technology and Research (A*STAR)\\[2pt]
    $^4$HSBC Business School, Peking University\\[7pt]
    \footnotesize
    \texttt{jinshuang.7@foxmail.com},\quad \texttt{bill.cjb@nus.edu.sg},\quad \texttt{zhang.guoyun@u.nus.edu}\\[2pt]
    \texttt{yzhao@phbs.pku.edu.cn},\quad \texttt{junyuz@nus.edu.sg}\\[2pt]
    \texttt{guanyi.w@nus.edu.sg},\quad \texttt{hzqin@nus.edu.sg}\\[6pt]
    $^*$Corresponding authors: Guanyi Wang and Hanzhang Qin.
  }
}
\date{}
\makeatletter
\renewcommand{\@maketitle}{%
  \newpage\null
  \begin{center}
    {\@title\par}\vskip 1em
    {\begin{tabular}[t]{@{}c@{}}\@author\end{tabular}\par}
  \end{center}
  \par\vskip 0.5em
}
\makeatother

\begin{document}
\raggedbottom
\maketitle
\begin{abstract}
Robust optimization (RO) and distributionally robust optimization (DRO) help decision makers manage uncertainty, but their practical use requires mathematical reformulation and implementation. Recent systems use large language models (LLMs) to generate code while relying on specialized modeling software to construct these reformulations. Their workflows offer limited support for user interaction, expertise reuse, and reformulation checking. To address these limitations, we present RobustOPT, an LLM-native human--AI collaborative modeling system for RO and DRO. LLMs directly derive explicit reformulations and generate code for standard optimization solvers, without relying on specialized RO/DRO modeling software. Users participate throughout the workflow by clarifying problem assumptions before reformulation, correcting models and customizing memory during reformulation, and revising decisions after solving. A customizable memory module enables users to maintain reusable expert knowledge and independently select Math memory for reformulation and Code memory for implementation. A pointwise reformulation check compares feasibility and worst-case cost at the same decision without requiring a known optimum. Application cases illustrate how uncertainty choices affect decisions and how users reoptimize after partial implementation. Experiments show that the memory module improves evaluation pass rates, while pointwise checks reveal reformulation errors hidden by agreement with a reference optimal value.
\end{abstract}

\input{sections/01_introduction}
\input{sections/02_related_work}
\input{sections/03_workflow}
\input{sections/04_memory}
\input{sections/05_check}
\input{sections/06_experiments}
\input{sections/07_conclusion}

\clearpage
\phantomsection\label{marker:references-start}
\bibliographystyle{plainnat}
\bibliography{references}

\clearpage
\phantomsection\label{marker:tables-start}
\input{tables/main_tables}

\clearpage
\phantomsection\label{marker:main-end}
\section*{Electronic Companion}
\addcontentsline{toc}{section}{Electronic Companion}
\setcounter{section}{0}
\renewcommand{\thesection}{EC.\arabic{section}}
\setcounter{equation}{0}
\renewcommand{\theequation}{EC.\arabic{equation}}
\setcounter{figure}{0}
\renewcommand{\thefigure}{EC.\arabic{figure}}
\setcounter{table}{0}
\renewcommand{\thetable}{EC.\arabic{table}}
\setcounter{proposition}{0}
\renewcommand{\theproposition}{EC.\arabic{proposition}}
\setcounter{example}{0}
\renewcommand{\theexample}{EC.\arabic{example}}
\setcounter{definition}{0}
\renewcommand{\thedefinition}{EC.\arabic{definition}}
\setcounter{lstlisting}{0}
\renewcommand{\thelstlisting}{EC.\arabic{lstlisting}}
\def\theHsection{EC.\arabic{section}}
\def\theHsubsection{EC.\arabic{section}.\arabic{subsection}}
\def\theHsubsubsection{EC.\arabic{section}.\arabic{subsection}.\arabic{subsubsection}}
\def\theHlstlisting{EC.\arabic{lstlisting}}
\def\theHlstnumber{EC.\arabic{lstlisting}.\arabic{lstnumber}}
\renewcommand{\theHequation}{EC.\arabic{equation}}
\renewcommand{\theHtable}{EC.\arabic{table}}
\renewcommand{\theHfigure}{EC.\arabic{figure}}
\input{appendices/companion}

\end{document}

%% file: sections/01_introduction.tex
\section{Introduction}\label{sec:intro}

Robust optimization (RO) and distributionally robust optimization (DRO) help decision makers choose actions when costs, demand, or supply are only partly known \citep{bertsimas2011theory,rahimian2022frameworks}. Consider a buyer allocating orders among suppliers with uncertain prices. Price ranges alone do not determine whether the buyer should protect against every supplier charging its highest price at once. The chosen treatment of uncertainty must be expressed in a model, preserved in its mathematical reformulation, and reflected in the computed allocation. When prices change or some orders have already been placed, the buyer must also revise the remaining decisions.


Formulating and solving this problem requires domain knowledge, mathematical derivation, and programming expertise. The decision maker specifies the operational requirements and the uncertainty to be covered. A modeler derives a \emph{finite reformulation}, a mathematical program with finitely many variables and constraints, and implements it in solver code. ROME and RSOME automate this transformation for models written in specialized modeling languages \citep{goh2011rome,chen2020rsome}. More recently, large language models (LLMs) have made natural-language descriptions a starting point for optimization modeling. ORLM and OptiMUS generate models and code \citep{huang2025orlm,ahmaditeshnizi2024optimus}, while DAOpt \citep{zhu2025daopt} and LLM4DRO \citep{wang2026llms} extend this approach to optimization under uncertainty. These two systems build on specialized modeling software such as RSOME, providing limited support for three practical needs: user interaction, expertise reuse, and reformulation checking.

First, practical modeling under uncertainty requires human oversight throughout the process. Initial descriptions may omit operating constraints or leave unclear which adverse conditions the decision must withstand. Automated derivations may then introduce errors that users need to identify and correct. After solving, prices or capacities may change while orders already placed must remain fixed. Interaction at individual stages helps with these needs, but users also need access to the intermediate information, equations, and code to follow the problem through successive revisions.

Second, effective use of expertise remains a challenge in LLM-based modeling. Established mathematical results and coding practices can help avoid errors that arise when an LLM relies on its own reasoning. Yet the knowledge needed to derive a reformulation differs from that needed to implement it. A duality rule may apply across solvers, whereas variable declarations and constraint interfaces depend on the chosen solver. Local interfaces and solver settings may also reflect users' requirements absent from the problem description. Users need relevant knowledge for both tasks and a way to adapt it to these requirements.


Third, users need immediate check on reformulation correctness during modeling, when the ground-truth optimum is unknown. Some existing systems provide code-execution feedback, but a program can run successfully and return a solution even when the reformulation omits a constraint or misrepresents the uncertainty. The challenge is to check for such discrepancies within the modeling workflow and provide evidence that helps users identify and correct errors, without the known optimum.


We present RobustOPT (\href{https://robustopt.tech}{\nolinkurl{robustopt.tech}}), an LLM-based modeling system for RO and DRO that supports human--AI collaboration throughout the modeling and solution process. Its LLM-native workflow uses LLMs to formulate the original model, derive an explicit finite reformulation, and generate code for standard optimization solvers, without relying on specialized RO/DRO modeling software.

\input{figures/fig_overview}

RobustOPT's workflow has six steps: Input, Identification, Original model, Reformulation, Solver code, and Results, as shown in Figure~\ref{fig:overview}. Starting from a description and data or an existing model, users confirm the modeling choices, review and revise the reformulation, and obtain solver code and a solution that they can subsequently update. Inputs, intermediate models, and code remain visible and editable throughout the process. \emph{Math memory} and \emph{Code memory} supply guidance for LLM-based derivation and implementation, respectively. Pointwise check provide evidence for revising the reformulation, and users can retain a reviewed correction in their own memory library. The three contributions are as follows.

\begin{enumerate}[leftmargin=*,label=\arabic*.]
\item \emph{An interactive workflow.} RobustOPT supports user interaction at three stages: before reformulation, during reformulation, and after solving. Before reformulation, users confirm facts and uncertainty assumptions. During reformulation, they correct the reformulation and customize the memory based on the feedback of pointwise reformulation check. After solving, they update data or fix selected decisions, then re-solve the revised problem, without another LLM call.
\item \emph{A customizable memory module.} Mathematical and coding knowledge are organized in independently selectable libraries. For each stage, users can choose system-provided knowledge, maintain their own entries, or proceed without memory. They can inspect the supplied rules and adapt them to their modeling and solver requirements.
\item \emph{A pointwise reformulation check.} For a given decision, an independent evaluator computes feasibility and worst-case cost from the original model and compares them with what the solver program reports. An adverse scenario, a distribution, or a pair of bounds can establish a discrepancy at that decision even when the optimal value is unknown.
\end{enumerate}

The application cases show how uncertainty choices and fixed decisions affect the solution, with all 40 complete-input tasks reaching independently evaluated optima. The system evaluation results show the memory module significantly improves the accuracy and efficiency. Generated programs and controlled model changes expose errors hidden by optimal-value agreement and show how checking feedback guides repair.

Section~\ref{sec:related} reviews related work. Sections~\ref{sec:workflow}--\ref{sec:check} present the workflow, memory module, and pointwise check. Section~\ref{sec:study} evaluates the three components, and Section~\ref{sec:conclusion} concludes. The electronic companion contains proofs, implementation details, experimental protocols, and additional results.

%% file: figures/fig_overview.tex
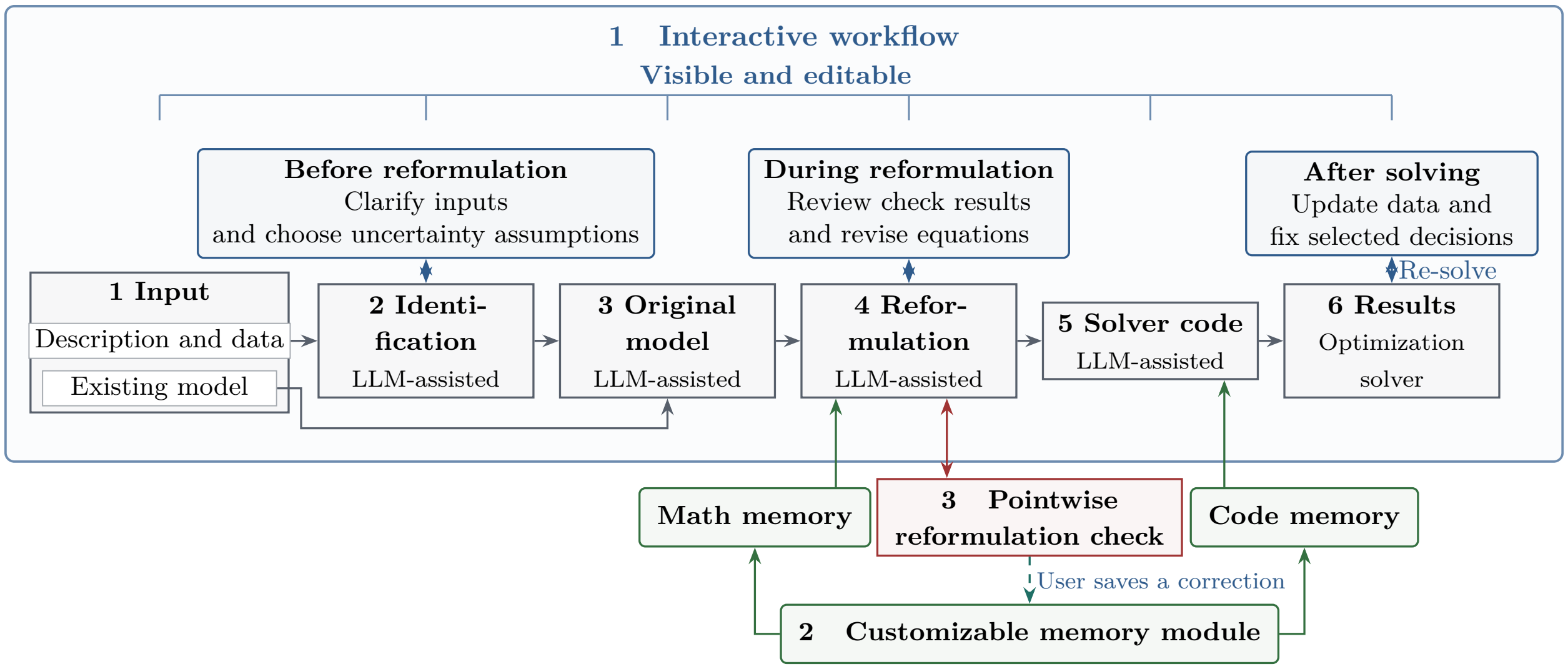
\begin{figure}[t]
\centering
\resizebox{\textwidth}{!}{%
\begin{tikzpicture}[
 font=\footnotesize, >={Stealth[length=2mm]},
 flowstep/.style={draw=rgray,fill=rgray!5,line width=.7pt,text width=2.37cm,
   minimum height=.88cm,align=center,inner xsep=3pt,font=\footnotesize\bfseries},
 phase/.style={draw=rblue,fill=rblue!5,line width=.7pt,rounded corners=2pt,
   minimum width=3.50cm,minimum height=1.02cm,align=center,inner xsep=5pt,font=\footnotesize},
 memory/.style={draw=rgreen,fill=rgreen!5,line width=.7pt,rounded corners=2pt,
   minimum height=.70cm,align=center,inner xsep=6pt,font=\footnotesize\bfseries},
 check/.style={draw=rred,fill=rred!5,line width=.7pt,minimum height=.70cm,
   align=center,inner xsep=6pt,font=\footnotesize\bfseries},
 flow/.style={->,draw=rgray,line width=.7pt},
 action/.style={<->,draw=rblue,line width=.7pt},
 guidance/.style={->,draw=rgreen,line width=.7pt},
]
\filldraw[draw=rblue!70,fill=rblue!2,line width=.8pt,rounded corners=3pt]
  (-1.85,-1.45) rectangle (16.85,4.02);
\node[text=rblue,font=\small\bfseries] at (7.5,3.67)
  {1\quad Interactive workflow};

\draw[rgray,fill=rgray!5,line width=.7pt]
  (-1.55,-.86) rectangle (1.55,.82);
\node[font=\footnotesize\bfseries] at (0,.57) {1 Input};
\node[draw=rgray!55,fill=white,minimum width=2.81cm,
  align=center,inner sep=2pt] (description) at (0,0)
  {Description and data};
\node[draw=rgray!55,fill=white,minimum width=2.81cm,
  align=center,inner sep=2pt] (model) at (0,-.57)
  {Existing model};

\node[flowstep] (s2) at (3.2,0)
  {2 Identification\\{\normalfont\scriptsize LLM-assisted}};
\node[flowstep] (s3) at (6.1,0)
  {3 Original model\\{\normalfont\scriptsize LLM-assisted}};
\node[flowstep] (s4) at (9.0,0)
  {4 Reformulation\\{\normalfont\scriptsize LLM-assisted}};
\node[flowstep] (s5) at (11.9,0)
  {5 Solver code\\{\normalfont\scriptsize LLM-assisted}};
\node[flowstep] (s6) at (14.8,0)
  {6 Results\\{\normalfont\scriptsize Optimization solver}};

\draw[flow] (description.east) -- (s2.west);
\draw[flow] (model.east) -- (1.70,-.57) -- (1.70,-1.05)
  -- (6.1,-1.05) -- (s3.south);
\foreach \i/\j in {2/3,3/4,4/5,5/6}{
  \draw[flow] (s\i.east) -- (s\j.west);
}

\draw[rblue!70,line width=.65pt] (0,2.95) -- (14.8,2.95);
\foreach \x in {0,3.2,6.1,9.0,11.9,14.8}{
  \draw[rblue!70,line width=.65pt] (\x,2.95) -- (\x,2.65);
}
\node[above,text=rblue,font=\footnotesize\bfseries] at (7.4,2.95)
  {Visible and editable};

\node[phase] (before) at (3.2,1.65)
  {\textbf{Before reformulation}\\
   Clarify inputs\\
   and choose uncertainty assumptions};
\node[phase] (during) at (9.0,1.65)
  {\textbf{During reformulation}\\
   Review check results\\
   and revise equations};
\node[phase] (after) at (14.8,1.65)
  {\textbf{After solving}\\
   Update data and\\
   fix selected decisions};

\draw[action] (before.south) -- (s2.north);
\draw[action] (during.south) -- (s4.north);

\draw[action] (after.south) --
  node[right,inner sep=2pt,text=rblue,font=\footnotesize] {Re-solve}
  (s6.north);

\node[memory] (math) at (7.15,-2.12) {Math memory};
\node[check] (check) at (10.45,-2.12)
  {3\quad Pointwise\\reformulation check};
\node[memory] (code) at (13.75,-2.12) {Code memory};

\coordinate (mathlink) at ($(math.east)!0.5!(s4.west)$);
\coordinate (checklink) at ($(check.west)!0.5!(s4.east)$);
\coordinate (codelink) at ($(code.west)!0.5!(s5.east)$);

\draw[guidance]
  (mathlink |- math.north) -- (mathlink |- s4.south);
\draw[<->,rred,line width=.7pt]
  (checklink |- s4.south) -- (checklink |- check.north);
\draw[guidance]
  (codelink |- code.north) -- (codelink |- s5.south);

\node[memory] (library) at (10.45,-3.52)
  {2\quad Customizable memory module};
\draw[guidance] (library.west) -| (math.south);
\draw[guidance] (library.east) -| (code.south);
\draw[->,dashed,rteal,line width=.7pt] (check.south) --
  node[right,inner sep=2pt,text=rblue,font=\scriptsize]
  {User saves a correction}
  (library.north);
\end{tikzpicture}}
\caption{Overview of the RobustOPT system.}
\label{fig:overview}\label{fig:workflow}
\end{figure}

%% file: sections/02_related_work.tex
\section{Related Literature}\label{sec:related}

Optimization software typically accepts a model written in a modeling language, whereas LLM-based systems can begin with a natural language-described problem. We review these systems through the three concerns that motivate RobustOPT: user interaction, customizable memory, and reformulation checking.

\subsection{Modeling Software and LLM-Based Systems}\label{sec:rel-systems}

General-purpose algebraic modeling languages such as JuMP provide a structured interface between mathematical models and optimization solvers \citep{lubin2023jump}. Specialized packages including ROME, YALMIP, JuMPeR, ROmodel, PyROS, and RSOME automate parts of robust or distributionally robust modeling \citep{goh2011rome,lofberg2012automatic,dunning2016advances,wiebe2022romodel,sherman2026pyros,chen2020rsome,chen2023rsome}. They draw on the theory of uncertainty sets, ambiguity sets, and tractable counterparts \citep{bental2009robust,bertsimas2004price,delage2010distributionally,wiesemann2014distributionally,mohajerin2018data,bental2015deriving}. The applicability conditions of a reformulation are essential: for example, \citet{xie2020tractable} gives sufficient conditions for tractable convex reformulations of two-stage DRO over a type-$\infty$ Wasserstein ball. Once the user has written the uncertain model, the software can apply a prescribed transformation or solution algorithm. RSOME, for example, can display the resulting standard-form program. Cutting-set methods use a worst-case routine at successive decisions to solve a robust problem \citep{mutapcic2009cutting,bertsimas2016reformulation}.

LLM-based systems also address the translation from a problem description to a model. NL4Opt studies the extraction of structured modeling information, while Chain-of-Experts, LLMOPT, ORLM, and OptiMUS generate deterministic models and solver programs \citep{ramamonjison2022nl4opt,xiao2024chain,jiang2025llmopt,huang2025orlm,ahmaditeshnizi2024optimus}. For optimization under uncertainty, DAOpt identifies uncertain quantities from a description and data, generates RSOME code with few-shot guidance, and revises the code using execution feedback \citep{zhu2025daopt}. LLM4DRO \citep{wang2026llms} presents an agentic workflow that validates an intermediate representation of the original DRO model against the user's description and translates it into RSOME code.

DAOpt and LLM4DRO have explored LLM-based modeling for RO and DRO through specialized modeling software such as RSOME. However, their workflows provide limited support for user interaction, expertise reuse, and reformulation checking. These limitations restrict users' ability to adapt and assess the generated models in practice. RobustOPT addresses these limitations through an LLM-native workflow that directly employs LLMs to derive explicit reformulations and generate code for standard solvers, without relying on specialized RO/DRO modeling software. RobustOPT supports user interaction throughout the modeling process, user customization and reuse of expert knowledge, and model-level reformulation checking, as shown in Table~\ref{tab:related}.



\subsection{Human Involvement in Optimization Modeling}\label{sec:rel-human}

The choice of a protection criterion is itself a modeling decision. \citet{iancu2014pareto} show that a worst-case optimal solution can be Pareto dominated and develop methods to obtain Pareto robustly optimal solutions, motivating scrutiny of performance beyond a single worst-case value. Robust satisficing evaluates fragility relative to a performance target \citep{long2023satisficing}, and its predictive extension accounts for both distributional ambiguity and parameter estimation uncertainty \citep{sim2025analytics}. Joint estimation and robustness optimization integrates parameter estimation with the choice of an uncertainty set that preserves desired performance \citep{zhu2022joint}. These approaches motivate making protection assumptions explicit to the user; they are distinct from the fixed-set RO and DRO counterparts evaluated in our experiments.

Interviews with optimization practitioners document iterative modeling workflows in which stakeholder dialogue, data preparation, and validation remain central \citep{lawless2025magical}. This evidence motivates supporting revision and review throughout modeling. Model management in decision support separates a model from its data and solver so that each can be reviewed and changed \citep{geoffrion1987introduction}. Mixed-initiative systems divide work between a user and an automated agent according to who can resolve each decision \citep{horvitz1999principles,amershi2019guidelines}. Interactive multiobjective optimization keeps the decision maker involved as preferences are formed \citep{miettinen2008introduction}. Among LLM systems, OptiMUS-0.3 lets users correct parameters, natural-language clauses, mathematical formulations, and code \citep{ahmaditeshnizi2026optimus}. OptiGuide answers what-if questions about a solved supply-chain model \citep{li2023large}. In DRO, \citet{wang2026llms} request clarification of missing modeling information.

RobustOPT connects confirmation of uncertainty assumptions, inspection and correction of the explicit reformulation, and revision of the computed decisions within one interactive workflow. When prices or capacities change, users can update the solution while holding decisions already implemented fixed.

\subsection{Memory for LLMs and for Optimization Modeling}\label{sec:rel-memory}

Reusable guidance can come from reference material or from experience. Retrieval-augmented generation supplies relevant documents \citep{lewis2020retrieval}, and MemGPT manages external memory \citep{packer2023memgpt}. Reflexion and ExpeL retain feedback or extract experience from earlier attempts \citep{shinn2023reflexion,zhao2024expel}. ReasoningBank and ACE distill strategies from past trajectories \citep{ouyang2026reasoningbank,zhang2025ace}. In optimization modeling, AlphaOPT builds an experience library of modeling insights, each with an applicability condition that can be refined across tasks \citep{kong2026alphaopt}.

For RO and DRO, \citet{bertsimas2025robust} provide expert informational prompts for robust and adaptive robust optimization. The library of \citet{wang2026llms} includes DRO formulation examples, RSOME implementation practices, and recurring errors and their repairs. Their library combines LLM reflection with human review. AutoREM learns reformulation rules for box, budgeted, and polyhedral RO through offline adaptation, using structured memory updates and validation-based acceptance \citep{chen2026autorem}.

These studies show how documents, expert rules, and experience can supply modeling knowledge. RobustOPT organizes such knowledge in a customizable memory module. Users can maintain their own entries and independently select, edit, or omit Math memory for reformulation and Code memory for implementation. A modeler can thus retain a mathematical rule while adapting the code to local solver conventions.

\parhead{Relation to AutoREM.} AutoREM and RobustOPT share memory-guided LLM reformulation and a subsequent coding stage. AutoREM learns mathematical experience from solver feedback and freezes the memory for online use. Its Appendix~G already distinguishes objective agreement from mathematical equivalence and reports expert audits of counterparts and coder fidelity. RobustOPT develops a complementary system contribution: users review uncertainty assumptions, formulations, code, and revised solutions; choose Math and Code libraries independently; and invoke automated comparisons of original and reformulated feasibility and cost at common decisions. These runtime comparisons do not require a known optimum. The analysis specifies premises for reformulation reuse and evidence that refutes exactness, while the experiments cover polyhedral and ellipsoidal RO and moment and Wasserstein DRO. The system-provided entries are written from the literature; they are not a new offline memory-learning algorithm. Finite pointwise checks can expose errors but cannot certify global equivalence. The two papers use different evaluation protocols, so their reported success rates are not directly comparable.

\subsection{Verification of Generated Models}\label{sec:rel-verify}

Optimization certificates and reformulation checks answer different questions. For example, the L0BnB framework obtains dual bounds and certifiably optimal solutions for a specified sparse regression model \citep{hazimeh2022sparse}; such a certificate presupposes the model being solved. RobustOPT checks whether a generated counterpart preserves the original uncertain model. Software validation offers several ideas for checking a generated reformulation. Translation validation checks an individual output of a transformation rather than establishing that every output will be correct \citep{pnueli1998translation}. Certifying algorithms supply independently checkable evidence \citep{mcconnell2011certifying}; an adverse scenario plays this role when it disproves a claimed robust guarantee. Property-based and metamorphic testing motivate examining multiple decisions and related parameter settings \citep{claessen2000quickcheck,chen2018metamorphic}. Among approaches for optimization models, FLARE proves MILP reformulations equivalent to a reference with a theorem prover \citep{robbins2026flare}, and ModelEquivBench compares generated and reference models across several relations \citep{zhu2026modelequivbench}. \citet{li2026falsification} test necessary conditions on transformed LP instances. OptArgus uses LLM auditors to find errors \citep{li2026optargus,wu2026fixwhatswrong}. As \citet{zhang2025optimind} observe, a model can reproduce a reference optimum and still be wrong.

RobustOPT's pointwise reformulation check asks whether the original model and its reformulation assign the same feasibility and worst-case cost to a given decision. Matching the reported optimum leaves that question open at other decisions. The check fixes the original decisions, optimizes over any auxiliary variables in the reformulation, and compares the result with an independent worst-case evaluation. It assesses whether the transformation preserves the original model, whose agreement with the application still requires modeling judgment.

%% file: sections/03_workflow.tex
\section{Interactive Workflow}\label{sec:workflow}

A user of RobustOPT can examine how a choice about uncertainty enters the model, review its mathematical consequences, and revise a solution when circumstances change. Here, a user may be a decision maker specifying operational requirements or a modeler reviewing the mathematics and code. A portfolio example illustrates these decisions over the six steps of the workflow. The online system also provides a step-by-step user manual.

\subsection{Overview}\label{sec:overview}

Users can begin with a problem description and data or with an existing uncertain model (Figure~\ref{fig:workflow}). LLMs extract the modeling information, draft the original model, derive its reformulation, and generate solver code. Users contribute domain knowledge and preferences, exercise modeling judgment, and decide whether to accept the proposed model and its revisions. They can supply missing information and inspect or edit the equations and code. Independent numerical routines perform the pointwise comparisons, and optimization solvers compute the solutions.

For a description, Identification extracts the decisions, constraints, parameters, and uncertain quantities and asks the user to resolve open questions. Original model then presents the proposed equations for review. A user who supplies an existing model can proceed directly to reviewing it. Once reviewed and accepted, the model serves as the working specification for reformulation and checking.

At Reformulation, the user can select mathematical rules from \emph{Math memory} and inspect the resulting equations alongside the available check results. These results help the user decide whether to continue, request another derivation, or edit the equations. At Solver code, the user can select the solver and implementation instructions from \emph{Code memory}, then review or revise the program. At Results, the user sees the solver status and, when a solution is available, its objective value and decisions. A fresh check can be requested if the program has been modified.

Users remain involved throughout the workflow. They clarify operational information \emph{before reformulation}, review and revise equations \emph{during reformulation}, and update data or fix implemented decisions \emph{after solving}.

\subsection{Before Reformulation: Choosing What to Protect Against}\label{sec:before}

Some questions at Identification concern facts the user can supply, such as a purchasing budget or a missing item in a data array. Others call for a modeling decision even when the observations are complete. Consider a manufacturer buying 20 interchangeable parts from four suppliers before prices are settled. The price range for each supplier still leaves open which prices may reach their upper limits together.

One uncertainty set allows all four suppliers to charge their highest prices at once. Another retains each supplier's bound but limits the combined size of the increases. By excluding some joint extremes, the latter may reduce the optimal worst-case cost. The manufacturer's \emph{protection choice} determines which joint price increases its purchases must withstand. At Identification the user reviews the extracted data and chooses among the alternatives. At Original model the user confirms that the objective, constraints, and uncertainty set express that choice.

\subsection{During Reformulation: Reviewing and Revising}\label{sec:during}

At Step~4, the LLM can review the proposed equations, while the \emph{pointwise check} provides an independent numerical comparison with the original model. The check asks whether the original model and the reformulation agree on feasibility and worst-case cost or return at selected decisions, without requiring a reference optimum. The report shows the available comparisons and their results. An adverse scenario or distribution can help a modeler understand a disagreement and locate its source. Section~\ref{sec:check} explains the checking mechanism.

A modeler who wants a new derivation can select \emph{Retry}. The LLM receives the previous reformulation and the report, together with any instructions the modeler adds, such as a reminder to preserve a stated mean equality. With \emph{Edit}, the modeler can instead change the displayed equations and check the revised formulation. The same option is available when the modeler prefers another representation, even if the report found agreement. A correction that applies more broadly can be saved in a personal memory library for future tasks.

\subsection{After Solving: Adjusting and Re-Solving}\label{sec:after}

A price or capacity may change after the original solve, while some orders have already been placed and cannot be changed. At Results the user can update available parameters, including the level of protection, fix the decisions already implemented, and reoptimize the remaining ones. When the changes are covered by the existing reformulation and can be applied to the program, re-solving requires no further LLM call.

Before re-solving, RobustOPT checks basic conditions on the revised data, such as a nonnegative uncertainty radius. The modeler reviews any additional assumptions required by the selected reformulation. The updated decisions and objective value can then be compared with the original solution. If the change calls for a new model or reformulation, the user can copy the task and return to the relevant earlier step.

\subsection{A Walkthrough}\label{sec:walkthrough}

In the two-asset portfolio from the RobustOPT gallery, an investor must allocate capital before either asset's return is known. The investor uses System-provided memory and later considers a revised forecast and a position that must remain fixed.

\parhead{Step 1: Input.} The investor supplies the expected returns, the information available about return errors, and the restrictions on the allocation.

\paperbox{Input: description and data}{%
\emph{Description.} Allocate up to one unit of capital between two assets before their returns are known. Borrowing and short sales are not allowed. Choose the allocation with the highest return we can guarantee for return errors in the range set by the research desk's shape matrix.\\[2pt]
\emph{Data.}\quad
\begin{tabular}{@{}lcc@{}}
 & asset 1 & asset 2\\
expected return & 10\% & 8\%\\
shape matrix $Q$ of return errors & $(0.0009,\ 0.0002)$ & $(0.0002,\ 0.0016)$
\end{tabular}}

\parhead{Step 2: Identification.} The capital constraint and nominal returns are specified. The remaining question is how the investor wishes to use the desk's shape matrix $Q$. The model can retain the joint ellipsoid described by $Q$ or use its componentwise ranges.

\paperbox{Identification: what the system found and what it asks}{%
\begin{tabular}{@{}ll@{}}
Decisions & shares $x_1,x_2\ge 0$, $x_1+x_2\le 1$ \\
Objective & maximize the guaranteed return \\
Uncertain & return errors $\delta$, centered at zero, shape matrix $Q$
\end{tabular}\\[3pt]
\emph{How should return errors be protected against?}\\
$\square$\ \ Componentwise intervals of $\pm3$ and $\pm4$ percentage points. Both errors may reach their worst values at once.\\
$\boxtimes$\ \ Joint ellipsoid $\delta^\top Q^{-1}\delta\le1$. The two errors cannot reach their worst values together.}

Figure~\ref{fig:portfolio} illustrates what this choice implies for a portfolio invested equally in both assets. The dot on the ellipse marks the worst pair of errors under joint protection, a loss of 2.69 percentage points. Using componentwise intervals admits the lower corner of the rectangle, marked by the square, where the loss reaches 3.50 percentage points.

\input{figures/fig_portfolio}

\parhead{Step 3: Original model.} Having chosen the ellipsoid, the investor reviews the objective and constraints in the displayed model. Each term is editable before confirmation.

\paperbox{Original model: confirm or edit}{%
\[
\max_{\substack{x\ge0\\x_1+x_2\le1}}\ \min_{\delta^\top Q^{-1}\delta\le1}\ (c+\delta)^\top x,\qquad c=(0.10,\ 0.08)^\top .
\]
\emph{Choice.} Confirm the joint ellipsoid and this model.}

\parhead{Step 4: Reformulation.} The investor selects System-provided Math memory. For a fixed allocation $x$, the largest loss from return errors in the ellipsoid is $\|L^\top x\|_2$, where $Q=LL^\top$. The reformulation bounds this loss by $t$, which is subtracted from the nominal return. The investor reviews the equations and the check report before proceeding.

\paperbox{Reformulation and check report}{%
\emph{Math memory.}\ \ $\boxtimes$ System-provided\ \ $\square$ User-customized\ \ $\square$ No memory\\[2pt]
\emph{Reformulation.} Factor the positive definite matrix $Q=LL^\top$. Maximize $c^\top x-t$ subject to $x_1+x_2\le1$, $x\ge0$, $w=L^\top x$, $\|w\|_2\le t$.\\[2pt]
\emph{Check report.} No discrepancy at the tested decisions.\\[2pt]
\emph{Next.}\ \ Continue, Retry, or Edit.}

\parhead{Step 5: Solver code.} The investor chooses Gurobi and System-provided Code memory. In the program below, the constraint defining $w_i$ uses column $i$ of $L$ to implement $L^\top x$. The investor can inspect or edit the program, or request new code before solving.

\paperbox{Solver code (excerpt)}{%
\emph{Solver.}\ Gurobi\qquad \emph{Code memory.}\ System-provided\\[2pt]
\texttt{L = np.linalg.cholesky(Q)}\\
\texttt{t = m.addVar(lb=0.0, name="t")}\\
\texttt{for i in range(n):}\\
\hspace*{1.6em}\texttt{m.addConstr(w[i] == gp.quicksum(L[j, i] * x[j] for j in range(n)))}\\
\texttt{m.addQConstr(gp.quicksum(w[i] * w[i] for i in range(n)) <= t * t)}\\
\texttt{m.setObjective(gp.quicksum(c[j] * x[j] for j in range(n)) - t, GRB.MAXIMIZE)}}

\parhead{Step 6: Results.} The investor sees the optimal allocation and its guaranteed return, with the models and check report available for inspection. Starting each time from that baseline, the investor considers two revisions. The first lowers asset~1's expected return to 9\%. The second retains the original forecasts and fixes $x_1=0.60$ to represent a position already established. Both revisions use the existing program and require no further LLM call.

\paperbox{Results and revised solutions}{%
\begin{tabular}{@{}lccl@{}}
 & asset 1 / asset 2 & guaranteed & request\\
Baseline & 94.0\% / 6.0\% & 7.01\% & --\\
Forecast revised & 79.3\% / 20.7\% & 6.15\% & change data: expected return of asset 1 to 9\%\\
Position fixed & 60.0\% / 40.0\% & 6.60\% & fix decision: $x_1=0.60$
\end{tabular}}

The lower forecast shifts 14.7 percentage points of capital to asset~2. Under the original forecasts, fixing the 60\% position in asset~1 reduces the guaranteed return by 0.41 percentage points. The investor can also change another parameter or fix further decisions and re-solve.

%% file: figures/fig_portfolio.tex
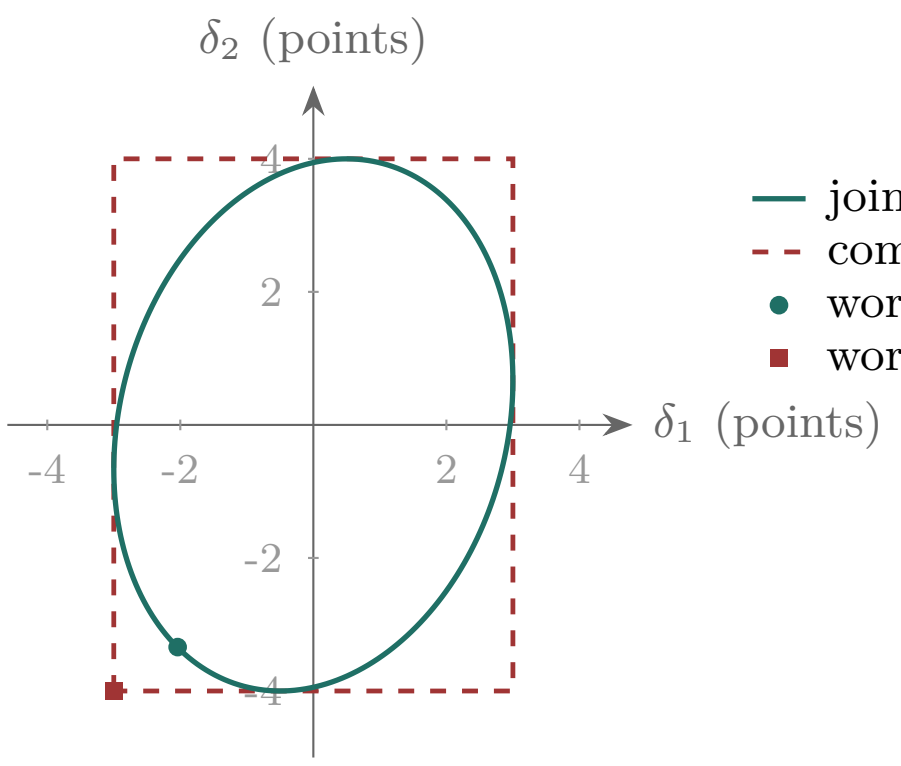
\begin{figure}[t]
\centering
\begin{tikzpicture}[>={Stealth[length=2mm]}, font=\footnotesize, scale=0.44]
\draw[->, black!60] (-4.6,0) -- (4.8,0) node[right] {$\delta_1$ (points)};
\draw[->, black!60] (0,-5.0) -- (0,5.1) node[above] {$\delta_2$ (points)};
\foreach \t in {-4,-2,2,4}{\draw[black!40] (\t,0.08) -- (\t,-0.08) node[below=1pt, font=\scriptsize] {\t};}
\foreach \t in {-4,-2,2,4}{\draw[black!40] (0.08,\t) -- (-0.08,\t) node[left=1pt, font=\scriptsize] {\t};}
\draw[rred, dashed, line width=0.9pt] (-3,-4) rectangle (3,4);
\draw[rteal, line width=1.0pt, domain=0:360, samples=120, smooth, variable=\t] plot ({3*cos(\t)}, {0.667*cos(\t)+3.944*sin(\t)});
\fill[rred] (-3,-4) +(-0.13,-0.13) rectangle +(0.13,0.13);
\fill[rteal] (-2.04,-3.34) circle (0.14);
\draw[rteal, line width=1.0pt] (6.6,3.4) -- (7.4,3.4) node[right, text=black] {joint ellipsoid};
\draw[rred, dashed, line width=0.9pt] (6.6,2.6) -- (7.4,2.6) node[right, text=black] {componentwise intervals};
\fill[rteal] (7.0,1.8) circle (0.14); \node[right] at (7.4,1.8) {worst for equal shares: loss 2.69};
\fill[rred] (7.0,1.0) +(-0.13,-0.13) rectangle +(0.13,0.13); \node[right] at (7.4,1.0) {worst for equal shares: loss 3.50};
\end{tikzpicture}
\caption{Joint and componentwise protection for the portfolio example. Deviations and losses are in percentage points.}
\label{fig:portfolio}
\end{figure}

%% file: sections/04_memory.tex
\section{The Customizable Memory Module}\label{sec:memory}

Deriving a finite reformulation requires mathematical rules for the uncertainty or ambiguity set at hand. Implementing the result also requires conventions specific to the solver. RobustOPT's \emph{memory module} lets users choose which rules the LLM receives for the derivation and which instructions it receives for the code. Users can adapt either to their own modeling practice, including interfaces to an existing code library, team conventions, and solver parameter settings. These local requirements need to be supplied explicitly alongside general mathematical knowledge.

\subsection{Organizing and Using Memory}\label{sec:module}

A \emph{memory library} is an editable collection of rules supplied to the LLM at the relevant stage. Each \emph{memory entry} groups related instructions and identifies their source and scope. At Step~4, \emph{Math memory} supplies solver-independent rules for deriving a finite reformulation. At Step~5, a \emph{Code memory} entry can give instructions specific to the selected solver, as the system-provided entries do for Gurobi.

Table~\ref{tab:entry} illustrates the distinction for an uncertain term $\delta^\top v$, where $v$ is its coefficient vector. In the portfolio model, $v$ is the allocation $x$. The mathematical rule gives the norm of $L^\top v$ for the worst deviation. The corresponding Gurobi instructions specify which elements of $L$ to use and which bounds to give the variables in the norm constraint.

Users select libraries independently for mathematical derivation and for the solver implementation reviewed at Step~5. The options are \emph{System-provided}, \emph{User-customized}, and \emph{No memory}. When the user selects a library, the module supplies the entries for the stage and the problem family, together with entries shared across families. The task records the library and entries supplied to the LLM. With No memory, generation proceeds without additional rules at that stage.

\subsection{System-Provided Memory}\label{sec:default}

The memory format accommodates rules for different RO and DRO formulations. The system-provided library used in the experiments contains four Math entries and four Gurobi Code entries, covering polyhedral RO, ellipsoidal RO, moment DRO, and Wasserstein DRO. Users can extend their libraries with entries for other formulations. The mathematical rules draw on work on polyhedral and ellipsoidal RO and uncertainty budgets \citep{bental2009robust,bertsimas2004price}, on ambiguity sets defined by moments \citep{delage2010distributionally}, and on type-1 Wasserstein balls \citep{mohajerin2018data}. The rules specify details that can change the model, such as which uncertain constraints need separate multipliers or how restrictions on the support enter a Wasserstein dual.

The Code entries translate these results into Gurobi variables and constraints. For example, a dual multiplier for an equality constraint in a polyhedral uncertainty set is free in sign. A new Gurobi variable is nonnegative by default, so the Code entry specifies an explicit lower bound of $-\infty$ for that multiplier.

\subsection{User-Customized Memory}\label{sec:maintain}

A modeler can inspect the system-provided entries and copy them to a personal library (Figure~\ref{fig:customize}). The modeler can then add a rule, change the conditions under which it applies, or remove an entry. The account retains the library for later tasks and keeps earlier revisions available.

\input{figures/fig_customize}

For example, a team may use the standard mathematical derivation but require its Gurobi models to have a prescribed name and to be built by a reusable function. A modeler can record these conventions in a personal Code entry and choose that library for implementation while keeping System-provided Math memory for the derivation. A check report may likewise reveal a missing bound that matters in other tasks. The modeler can review the correction and its scope, save it as an entry, and inspect subsequent formulations and programs for consistency with the saved rule.

\FloatBarrier

%% file: figures/fig_customize.tex
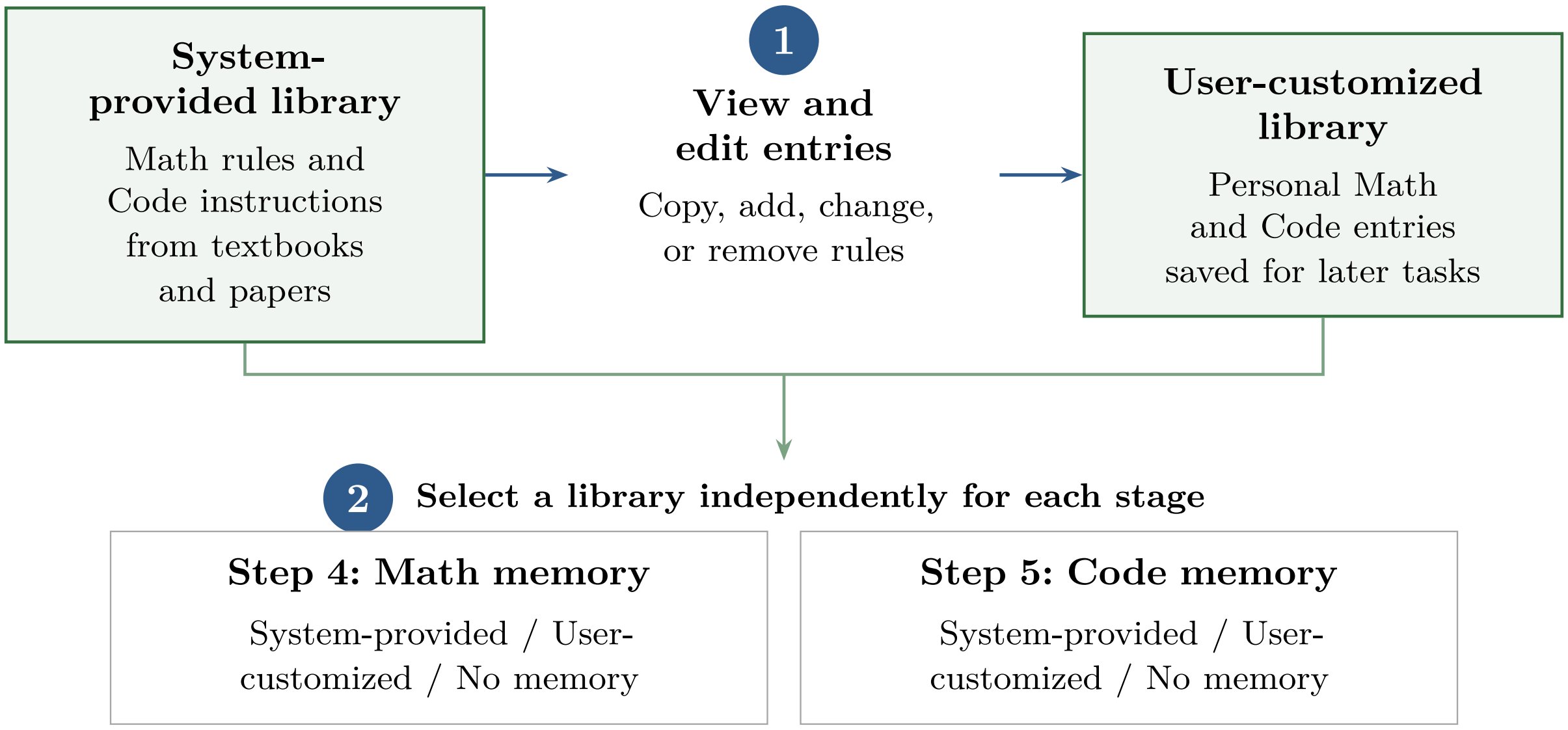
\begin{figure}[!htbp]
\centering
\resizebox{.96\textwidth}{!}{%
\begin{tikzpicture}[
 font=\small, >={Stealth[length=2mm]},
 library/.style={draw=rgreen,line width=.8pt,fill=rgreen!7,align=center,inner sep=9pt,text width=3.8cm,minimum height=1.65cm},
 action/.style={align=center,text width=3.6cm},
 badge/.style={circle,draw=rblue,fill=rblue,text=white,font=\bfseries,inner sep=3pt},
 choice/.style={draw=black!35,fill=white,align=center,text width=5.6cm,inner sep=7pt},
]
\draw[rgreen!55,fill=rgreen!3] (-2.08,.88) rectangle (2.02,-.77);
\draw[rgreen!55,fill=rgreen!3] (-2.02,.82) rectangle (2.08,-.83);
\node[library] (system) at (0,0) {\textbf{System-provided library}\\[4pt]\footnotesize Math rules and Code instructions\\from textbooks and papers};
\draw[rgreen!55,fill=rgreen!3] (7.92,.88) rectangle (12.02,-.77);
\draw[rgreen!55,fill=rgreen!3] (7.98,.82) rectangle (12.08,-.83);
\node[library] (personal) at (10,0) {\textbf{User-customized library}\\[4pt]\footnotesize Personal Math and Code entries\\saved for later tasks};
\node[badge] at (5,1.25) {1};
\node[action] at (5,0) {\textbf{View and edit entries}\\[3pt]\footnotesize Copy, add, change, or remove rules};
\draw[->,rblue,line width=.9pt] (system.east) -- (3,0);
\draw[->,rblue,line width=.9pt] (7,0) -- (personal.west);
\draw[rgreen!65,line width=.8pt] (system.south) -- (0,-1.85) -- (10,-1.85) -- (personal.south);
\draw[->,rgreen!65,line width=.8pt] (5,-1.85) -- (5,-2.65);
\node[badge] at (1.05,-3.0) {2};
\node[anchor=west,font=\footnotesize\bfseries] at (1.45,-3.0) {Select a library independently for each stage};
\node[choice] at (1.8,-4.20) {\textbf{Step 4: Math memory}\\[4pt]\footnotesize System-provided / User-customized / No memory};
\node[choice] at (8.2,-4.20) {\textbf{Step 5: Code memory}\\[4pt]\footnotesize System-provided / User-customized / No memory};
\end{tikzpicture}}
\caption{Selecting and maintaining Math and Code memory.}
\label{fig:customize}
\end{figure}

%% file: sections/05_check.tex
\section{Pointwise Reformulation Check}\label{sec:check}

The pointwise reformulation check compares a solver implementation of the candidate reformulation with an independent evaluation of the original model. At a fixed decision, it tests whether the two agree on feasibility and, when feasible, worst-case cost. This comparison can reveal an error even when the models agree on the optimal decision and its value.

\subsection{Reformulations and Silent Failures}\label{sec:prelim}

Once an order has been optimized, a buyer may want to consider a different split between suppliers. The first example shows how a reformulation can misprice that split. The second shows how it can accept an order that cannot be filled.

\begin{example}\label{ex:shared}
A buyer splits a unit order between two suppliers whose surcharges move in opposite directions. When supplier~1 charges $\xi$, supplier~2 charges $1-\xi$, with $\xi\in[0,1]$. Supplier~1 also charges a premium $a$ per unit. If $x\in[0,1]$ is the share bought from supplier~1, the worst-case cost over all distributions of $\xi$ on $[0,1]$ is
\[
F_0(x;a)=ax+\sup_{\Prob}\E_\Prob\big[x\xi+(1-x)(1-\xi)\big]=ax+\max\{x,1-x\}.
\]
A reformulation that maximizes each supplier's surcharge separately gives $g(x;a)=ax+x+(1-x)=ax+1$. This ignores the fact that both surcharges depend on the same $\xi$ and cannot attain their maxima together. At $a=2$, both models buy everything from supplier~2 at cost $1$. At an equal split the surcharges offset each other exactly, so the worst-case cost is $3/2$. The reformulation charges $2$. If the premium falls to zero, splitting the order reduces the true worst-case cost to $1/2$, while the reformulation reports $1$ for every split (Figure~\ref{fig:example}).
\end{example}

\input{figures/fig_example}

\begin{example}\label{ex:capacity}
A buyer needs ten units. Supplier~A charges one per unit and supplier~B charges two. Under adverse conditions B can deliver only four units. At these prices both the original model and a program that omits B's delivery limit buy everything from A at cost ten. Requiring six units from B exposes the missing constraint. The program accepts an order that B may be unable to fill.
\end{example}

Such errors can arise when a derivation uses an upper price endpoint without checking whether the purchase quantity can be negative, treats a fixed mean as an inequality bound, or evaluates costs exposed to the same shock under separate worst cases \citep{gorissen2015practical}. Their common feature is that the models disagree about at least one decision. To compare them, we need to identify that decision in both formulations.

\parhead{Comparing the models at a fixed decision.} Let $\theta$ collect the data, including costs, capacities, and the parameters of the uncertainty or ambiguity set. Write the original model as
\begin{equation}\label{eq:original}
P(\theta):\quad \min_{x\in\cD(\theta)}\; F_0(x;\theta)\quad\text{subject to}\quad F_i(x;\theta)\le 0,\quad i=1,\dots,m.
\end{equation}
The set $\cD(\theta)$ imposes the original decision domains, including integrality, and the deterministic constraints. For $i=0,\ldots,m$, the RO expression $F_i(x;\theta)=\sup_{\xi\in\cU(\theta)} f_i(x,\xi;\theta)$ is the worst case over the uncertainty set. For a DRO worst expectation, $F_i(x;\theta)=\sup_{\Prob\in\cP(\theta)}\E_\Prob f_i(x,\xi;\theta)$ is the worst expected value over the ambiguity set. A maximization problem takes the form \eqref{eq:original} on negating its objective. Let $\cX(\theta)$ denote the feasible set of $P(\theta)$. We combine its cost and feasibility in one function, the \emph{extended worst-case cost} $\Fbar(x;\theta)$, equal to $F_0(x;\theta)$ on $\cX(\theta)$ and $+\infty$ elsewhere in $\R^n$.

The candidate finite program is
\begin{equation}\label{eq:reform}
C(\theta):\quad \min_{z}\; G_0(z;\theta)\quad\text{subject to}\quad z\in\cG(\theta).
\end{equation}
Its variables $z$ also include auxiliary quantities such as dual multipliers. The \emph{decision map} $\varphi$ identifies the original decisions $x=\varphi(z)$ in the program. For an order split $\hat x$, for example, the split is fixed but the program must still choose the auxiliary variables. The least cost the program assigns to any given $x$ is its \emph{projected value}
\begin{equation}\label{eq:projected}
g(x;\theta)=\inf\{G_0(z;\theta): z\in\cG(\theta),\ \varphi(z)=x\},
\end{equation}
with $g=+\infty$ if the program has no feasible solution at $x$. The two quantities to compare are now $\Fbar(x;\theta)$ and $g(x;\theta)$. They express the extended cost assigned to the same decision in the original model and the finite program.

Table~\ref{tab:families} illustrates how the original model can be evaluated independently at a fixed decision. For a budgeted constraint one need only sort the exposures and allocate the budget to the largest. For the moment-DRO losses and convex supports in the table, the evaluator computes the worst expectation through a finite convex program over distributional masses and locations, subject to the original support and moment conditions. These calculations use the uncertainty description directly, without relying on the candidate reformulation. Section~\ref{ec:oracles} gives the derivations and conditions for the implemented routines.

\subsection{Exactness and Reformulation Records}\label{sec:records}

We define exactness by preservation of the extended worst-case cost. For decisions with finite worst-case cost, this also preserves feasibility. Some formulations are deliberately conservative and are intended to bound the cost from above.

\begin{definition}[Exactness and conservativeness]\label{def:exact}
A finite program $C(\theta)$ is \emph{exact} for $P(\theta)$ if $g(x;\theta)=\Fbar(x;\theta)$ for every $x\in\R^n$, and \emph{conservative} if $g(x;\theta)\ge\Fbar(x;\theta)$ for every $x\in\R^n$.
\end{definition}

Defining both functions on all of $\R^n$ also captures decisions that violate the original domains or constraints. Because an infeasible decision and a feasible decision with infinite worst-case cost both have extended value $+\infty$, the checker records constraint feasibility separately from cost. A conservative program may exclude some feasible decisions or charge them more than their original worst-case cost. The program of Example~\ref{ex:capacity} fails even the conservative requirement by accepting an infeasible order. The program in Example~\ref{ex:shared} is conservative. Its excess charge for a split order matters precisely because the buyer wanted an exact cost.

The same distinction matters on a later solve. A price may change, or an order that has already been placed may have to be held fixed. A reformulation record makes explicit the premises under which the same equations apply.

\begin{definition}[Reformulation record]\label{def:record}
A \emph{reformulation record} is a tuple $\cR=(P,\Pi,C,\varphi)$ of the original model, a finite set $\Pi$ of premises on $\theta$, the finite program, and the decision map. Its \emph{premise region} is $\ThetaPi=\{\theta:\pi(\theta)\text{ holds for all }\pi\in\Pi\}$. The record is \emph{valid} if $C(\theta)$ is exact for $P(\theta)$ for every $\theta\in\ThetaPi$.
\end{definition}

The next proposition states when the same reformulation can support a revised decision problem without a new derivation. Its premise is exactness throughout the stated parameter region.

\begin{proposition}[Reuse of a reformulation]\label{prop:record}
For a valid record, at every $\theta\in\ThetaPi$ the two models have the same optimal value, and each optimal program solution yields an optimal original decision. Exactness persists after any restriction on the original decisions alone, including fixing some of them. If a program is only conservative at the data used, each feasible program solution still yields a feasible original decision with worst-case cost no greater than its reported cost.
\end{proposition}

For example, an endpoint rule for interval prices can still be used at new prices if its premise of nonnegative purchase quantities continues to hold. Table~\ref{tab:evidence} distinguishes the derivation and premises needed over the whole region from the numerical agreement observed at particular decisions. Section~\ref{ec:proofs} gives the proof and the converse for attained projected values.

\subsection{The Multi-Level Check}\label{sec:levels}

At a selected decision $\hat x$, the checker fixes the original variables in the solver program and optimizes over its auxiliary variables. An independently written evaluator computes feasibility and worst-case cost at the same decision from a structured, machine-readable representation of the original model. This representation specifies the decisions, constraints, and uncertainty information required by the evaluator. The next proposition identifies the evidence needed to reject exactness. An admissible scenario or distribution can expose a cost understatement or an infeasible decision accepted by the program. To establish a cost overstatement, a lower bound on the program's projected value must exceed an upper bound on the original worst-case cost.

\par\addvspace{0.65\baselineskip}
\noindent\begin{minipage}{\linewidth}
\begin{proposition}[Pointwise refutation]\label{prop:counter}
At a given decision $\hat x$, either of the following refutes exactness.
\begin{enumerate}[label=(\alph*),leftmargin=*]
\item A feasible program solution $\hat z$ maps to $\hat x$. Either $\hat x$ violates its original domain, or an admissible RO scenario or a DRO distribution in the ambiguity set violates an original constraint or yields a cost above $G_0(\hat z;\theta)$. This also refutes conservativeness.
\item The original model accepts $\hat x$, and bounds on its cost and the projected program value satisfy $F_0(\hat x;\theta)\le u<\ell\le g(\hat x;\theta)$.
\end{enumerate}
\end{proposition}
\end{minipage}\par\addvspace{0.65\baselineskip}

In Example~\ref{ex:capacity}, ordering six units from the second supplier gives part~(a) when adverse conditions limit its delivery to four units. For an equal split in Example~\ref{ex:shared}, the bounds $u=3/2$ and $\ell=2$ give part~(b) at $a=2$. This overstatement refutes exactness and remains consistent with conservativeness. The proof is in Section~\ref{ec:proofs}.

The framework combines several levels of checking (Table~\ref{tab:levels}). Level M examines the original model's feasibility, S checks program execution, and D1 evaluates the returned decision. D2 also examines nearby decisions, vertices when practical, and random trials. An order split is one such decision at which the error in Example~\ref{ex:shared} becomes visible. For parameterized models and programs, \Dt{} repeats comparisons at changed data. D0 compares optimal values when a reference is available for benchmarking. Figure~\ref{fig:pointwise} shows the comparison shared by D1, D2, and \Dt{}, while Table~\ref{tab:levels} specifies the role of each level.

\input{figures/fig_pointwise}

In RobustOPT, Step~4 presents the reformulation with the available check results. For these numerical comparisons, the system prepares a program using its default solver and Code memory settings. At Step~5, users inspect the program and can customize the solver, Code memory, or code. If these choices change the implementation, a new comparison is needed to assess the revised program. Appendix~\ref{ec:schema-map} describes this implementation.

The check report identifies the decision behind each numerical disagreement, or \emph{flag}, and gives the two evaluations and any available supporting evidence. A flag becomes a \emph{counterexample} when a witness or separated bounds satisfy Proposition~\ref{prop:counter} at a common decision. Both evaluations must use the same decision after accounting for numerical tolerances. For example, a trial point just outside a variable bound may be moved to the bound when fixing the program variables (Section~\ref{ec:threshold-diagnostic}). One counterexample suffices to refute exactness. The user can review the implicated equations and code, edit them or request another derivation, and save a reusable correction in the memory library.

A completed check with no flag records agreement within tolerance at the decisions and data examined. The report distinguishes that outcome from an incomplete comparison or one for which no applicable worst-case routine exists. To locate a discrepancy, the modeler reviews the structured representation against the original equations and the program against the displayed reformulation. The disagreement may originate in either representation or in the derivation itself. Model review addresses whether the original equations express the user's requirements; the pointwise check examines their transformation into a finite program. Section~\ref{ec:checker} gives the sampling and numerical protocols.

%% file: figures/fig_example.tex
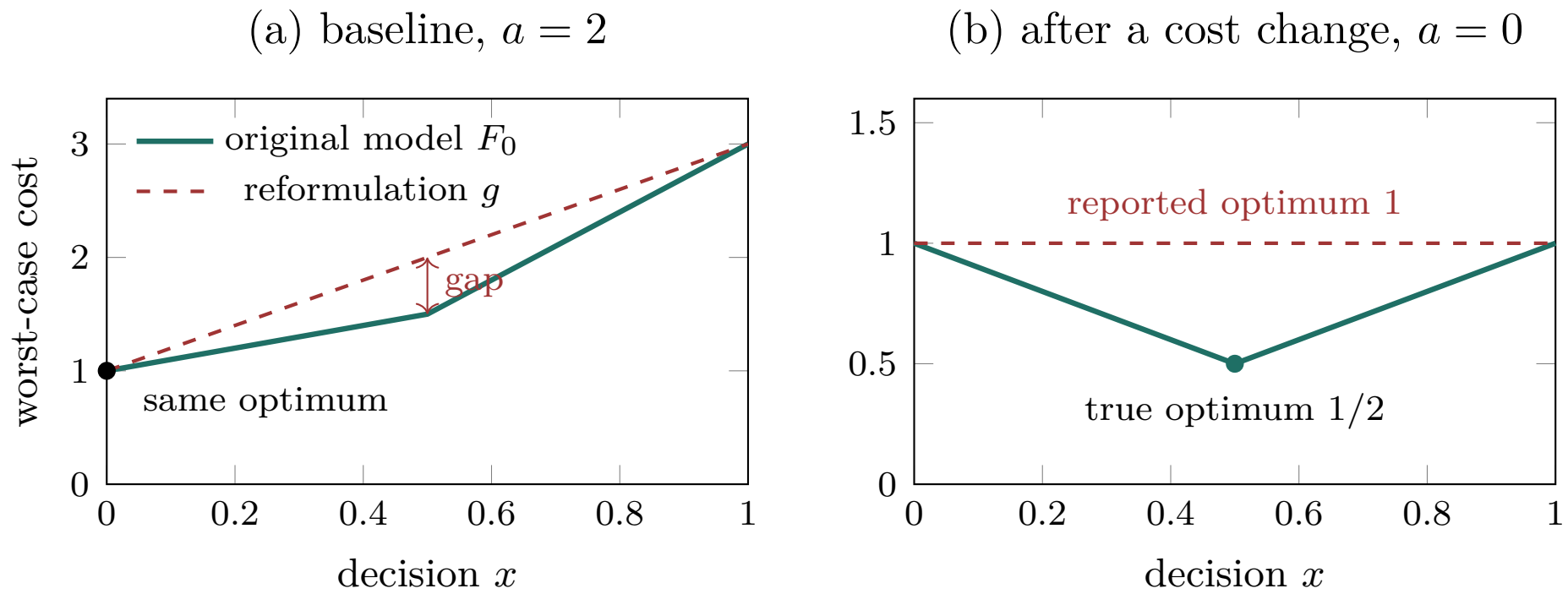
\begin{figure}[!hb]
\centering
\begin{tikzpicture}
\begin{axis}[name=a1, width=6.6cm, height=4.6cm, xmin=0, xmax=1, ymin=0, ymax=3.4,
  xlabel={decision $x$}, ylabel={worst-case cost}, title={\small (a) baseline, $a=2$},
  tick label style={font=\scriptsize}, label style={font=\footnotesize},
  legend style={font=\scriptsize, draw=none, fill=none, at={(0.02,0.98)}, anchor=north west}]
\addplot[rteal, very thick, domain=0:1, samples=101] {2*x + max(x,1-x)};
\addplot[rred, thick, dashed, domain=0:1] {2*x + 1};
\addplot[only marks, mark=*, mark size=1.8pt, black] coordinates {(0,1)};
\legend{original model $F_0$, reformulation $g$}
\node[font=\scriptsize, anchor=west] at (axis cs:0.03,0.72) {same optimum};
\draw[<->, rred] (axis cs:0.5,1.5) -- (axis cs:0.5,2) node[midway, right, font=\scriptsize] {gap};
\end{axis}
\begin{axis}[at={(a1.east)}, anchor=west, xshift=1.3cm, width=6.6cm, height=4.6cm, xmin=0, xmax=1, ymin=0, ymax=1.6,
  xlabel={decision $x$}, title={\small (b) after a cost change, $a=0$},
  tick label style={font=\scriptsize}, label style={font=\footnotesize}]
\addplot[rteal, very thick, domain=0:1, samples=101] {max(x,1-x)};
\addplot[rred, thick, dashed, domain=0:1] {1};
\addplot[only marks, mark=*, mark size=1.8pt, rteal] coordinates {(0.5,0.5)};
\node[font=\scriptsize, anchor=north] at (axis cs:0.5,0.45) {true optimum $1/2$};
\node[font=\scriptsize, anchor=south, text=rred] at (axis cs:0.5,1.02) {reported optimum $1$};
\end{axis}
\end{tikzpicture}
\caption{Optimal-value agreement and pointwise discrepancies (Example~\ref{ex:shared}).}
\label{fig:example}
\end{figure}

%% file: figures/fig_pointwise.tex
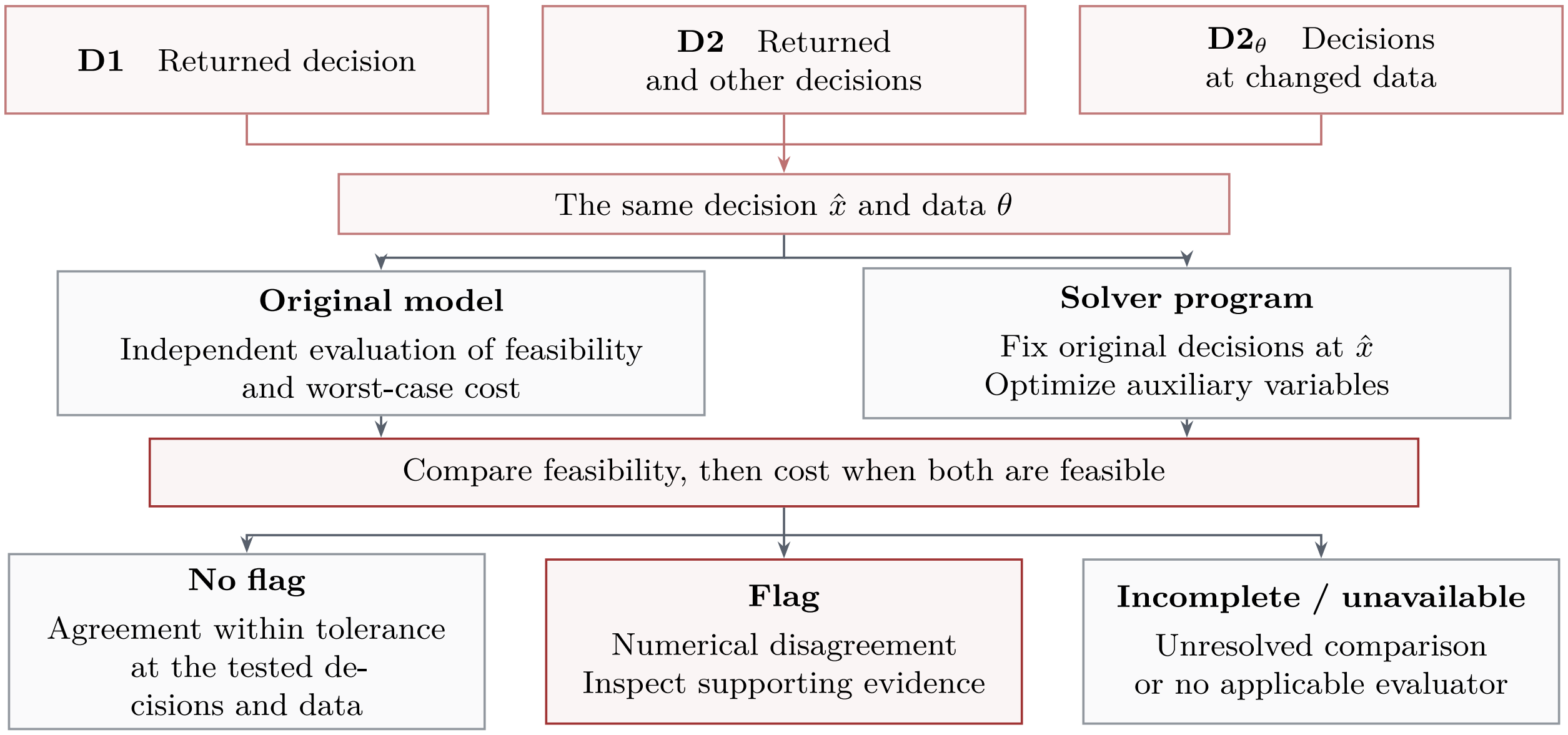
\begin{figure}[!htbp]
\centering
\resizebox{\textwidth}{!}{%
\begin{tikzpicture}[
 font=\fontsize{9}{11}\selectfont, >={Stealth[length=1.7mm]},
 panel/.style={draw=rgray!65,fill=rgray!3,line width=.65pt,align=center,inner sep=5pt},
 arr/.style={->,draw=rgray,line width=.65pt},
 level/.style={panel,draw=rred!65,fill=rred!4,text width=4.5cm,minimum height=1.08cm},
 outcome/.style={panel,text width=4.5cm,inner sep=4pt,minimum height=1.65cm},
]
\node[level] (d1) at (2.65,0) {\textbf{D1}\quad Returned decision};
\node[level] (d2) at (8.05,0) {\textbf{D2}\quad Returned and other decisions};
\node[level] (dt) at (13.45,0) {\textbf{\Dt}\quad Decisions at changed data};
\node[panel,draw=rred!65,fill=rred!4,text width=8.6cm,minimum height=.60cm] (point) at (8.05,-1.45)
 {The same decision $\hat x$ and data $\theta$};
\draw[rred!70,line width=.65pt] (d1.south) -- (2.65,-.85) -- (13.45,-.85) -- (dt.south);
\draw[->,rred!70,line width=.65pt] (d2.south) -- (point.north);
\node[panel,text width=6.15cm,minimum height=1.22cm] (original) at (4.0,-2.85)
 {\textbf{Original model}\\[3pt]Independent evaluation of feasibility\\and worst-case cost};
\node[panel,text width=6.15cm,minimum height=1.22cm] (program) at (12.1,-2.85)
 {\textbf{Solver program}\\[3pt]Fix original decisions at $\hat x$\\Optimize auxiliary variables};
\draw[arr] (point.south) -- (8.05,-1.99) -| (original.north);
\draw[arr] (8.05,-1.99) -| (program.north);
\node[panel,draw=rred,fill=rred!5,text width=12.4cm,minimum height=.68cm] (compare) at (8.05,-4.15)
 {Compare feasibility, then cost when both are feasible};
\draw[arr] (original.south) -- ($(compare.north)+(-4.05,0)$);
\draw[arr] (program.south) -- ($(compare.north)+(4.05,0)$);
\node[outcome] (agree) at (2.65,-5.85)
 {\textbf{No flag}\\[3pt]Agreement within tolerance\\at the tested decisions and data};
\node[outcome,draw=rred,fill=rred!5] (flag) at (8.05,-5.85)
 {\textbf{Flag}\\[3pt]Numerical disagreement\\Inspect supporting evidence};
\node[outcome] (missing) at (13.45,-5.85)
 {\textbf{Incomplete / unavailable}\\[3pt]Unresolved comparison\\or no applicable evaluator};
\draw[arr] (compare.south) -- (8.05,-4.78) -| (agree.north);
\draw[arr] (8.05,-4.78) -- (flag.north);
\draw[arr] (8.05,-4.78) -| (missing.north);
\end{tikzpicture}}
\caption{Pointwise comparison across checking levels.}
\label{fig:pointwise}
\end{figure}

%% file: sections/06_experiments.tex
\section{System Demonstration and Evaluation}\label{sec:study}

Our evaluation examines the workflow's support for clarifying uncertainty choices, editing displayed formulations, and fixing implemented decisions. We also compare Math and Code memory choices and examine whether pointwise comparisons reveal reformulation errors hidden by a correct optimal value. The constructed tasks cover polyhedral RO, ellipsoidal RO, moment DRO, and Wasserstein DRO, with benchmark reference values verified by independent numerical calculations. Appendix~\ref{ec:task-construction} describes their construction and reference checks.

\label{sec:setup}
The main experiments use DeepSeek-V4.1-Flash-1017 as the base LLM and Gurobi as the optimization solver. Workflow cases start from descriptions and data. Each workflow task allows up to three attempts. Within an attempt, automatic repair returns an execution error or check result to the LLM as feedback for revision, subject to the limits in Appendix~\ref{ec:suite}. The run stops once solving and all requested checks complete without a flag; the independent evaluator then assesses the delivered solution. Memory comparisons start from original models and allow one attempt per configuration and task. Appendix~\ref{ec:ledger} gives the settings and benchmarks.

Objective agreement ($\OA$) requires the reported optimum to lie within $10^{-5}(1+|f^\star|)$ of the reference value. An evaluation pass ($\EP$) additionally requires the returned decision to be feasible in the original model, its independently evaluated worst-case cost to agree with the reported value, and a completed pointwise check with no flag. Independent programs evaluate original feasibility and cost in the workflow cases. In the memory experiments the evaluator also compares the generated program with the benchmark at sampled decisions and at the reference optimizer, which only the evaluator sees.

Estimated token costs are reported in US dollars (USD). Mean elapsed times include unsuccessful attempts and repairs. The comparison across workflow models uses API-call time.

\subsection{Evaluating the Interactive Workflow}\label{sec:eval-workflow}

The workflow suite contains 40 tasks, ten per family, covering signed and integer decisions, fractional uncertainty budgets, different matrix conventions, moment bounds, finite supports, and instances with 50--80 decisions. We also examine four of these tasks with open uncertainty choices (N01--N04) and four with post-solve changes (N05--N08). Table~\ref{tab:suite} summarizes the runs.

The service allocation case (N02) illustrates why uncertainty choices matter. A service provider allocates 20 divisible blocks among five sites whose costs depend on common factors through a loading matrix $L$. The initial description leaves the joint range of the factors unspecified. The clarification specifies a Euclidean factor ball of radius 1.6. The resulting protection term is $1.6\|L^\top x\|_2$, and the optimal worst-case cost is CNY~8,389. Allowing each factor to reach the same bounds independently changes the protection term to $1.6\|L^\top x\|_1$. Under this box the original allocation costs CNY~10,336. Reallocating 7.77 of the 20 blocks reduces the cost to CNY~9,761.

With all modeling choices specified in the inputs, the base LLM at high reasoning and No memory delivered independently verified optimal solutions on the first attempt for all 40 tasks. None required code repair or check-guided repair. Supplementary results for GPT-6.1 Sol on the same inputs are reported in Appendix~\ref{ec:suite} (Table~\ref{tab:workflow-models}).

Whereas the 40-task batch supplied complete inputs, separate runs left the protection choices in N01--N04 open to test interactive clarification. All four prompted clarification. The inventory case (N03) also illustrates correction through the interface. The LLM review identified interchanged holding and shortage costs in the initial objective and stopped the workflow for correction. In a subsequent continuation, we edited the displayed objective to charge holding for surplus stock and shortage for unmet demand. The revised solution $x=(6,4,9,7)$ attained a worst-case cost of 42.125, confirmed by an independent distribution-based LP. Appendix~\ref{ec:suite} distinguishes these initial runs from the continuation after editing.

After solving, changes in prices or capacity may require a manager to revise the remaining decisions while keeping orders already placed fixed. In N05--N08 we first changed operating costs and then strengthened protection. The next two requests fixed successive groups of ten decisions at quantities specified in the original case baseline. Each of the 16 revised solutions matched an independently computed optimum for that particular combination of updated data and fixed decisions (Table~\ref{tab:reuse}). Nine requests updated inputs or variable bounds directly in the retained program. For the other seven, the generated code did not expose the affected quantity as an adjustable input, so RobustOPT generated a revised program in a copy of the task.

The last three columns of Table~\ref{tab:reuse} hold the updated data and protection level constant while fixing first one group of decisions and then a second. Their successive differences, calculated before rounding, give the increase in the optimal objective value from each additional group of fixed decisions. In procurement (N05), fixing the first group of orders adds CNY~67.53 million and fixing the second adds a further CNY~77.65 million. In freight (N08), the corresponding increases are CNY~23.57 thousand and CNY~37.48 thousand.

\subsection{Evaluating the Memory Module}\label{sec:eval-memory}

Using the base LLM with reasoning disabled, we first compared three memory configurations on 256 original models. With No memory, the generated programs passed evaluation on 81 tasks. A comparison library containing published RO rules \citep{bertsimas2025robust} and our DRO extensions yielded 188. With the four Math and four Code entries of the system-provided library, the count rose to 241 (Table~\ref{tab:memory-final4}). The models have two to five decisions in RO and two to eight in DRO. With system-provided guidance all 64 ellipsoidal cases and 62 of 64 in each DRO family passed. Polyhedral RO accounted for 11 of the 15 remaining cases.

\FloatBarrier

With system-provided memory and reasoning disabled, 241 of 256 tasks passed evaluation (94.14\%). With reasoning enabled and No memory, 255 passed (99.61\%), with one timeout. The memory-assisted configuration passed 14 fewer tasks while reducing total token cost from \$9.38 to \$2.90 and mean time per task from 340.4 to 41.7 seconds. These reductions amount to 69.1\% in cost and 87.8\% in elapsed time (Figure~\ref{fig:memcost}). Of the 242 programs that passed the runtime check with system-provided memory, 241 also met the independent evaluation criteria (Table~\ref{tab:runtime-outcomes}).

\input{figures/fig_memcost}

To study the effect of Math memory and Code memory, we enabled Math and Code entries separately for LLM on a set of 64 challenging models, 16 per family, with reasoning disabled. The RO models have five decisions and the DRO models have six to eight. Math entries alone led to evaluation passes on 50 tasks, close to the 55 obtained with both types. Code entries alone yielded 13, compared with 15 when neither was supplied (Table~\ref{tab:stage-selection}). The LLM received the Math entries both for the initial derivation and for revisions after a failed check. The stronger result for Math memory is consistent with the need for a correct reformulation before coding. Code entries guide its implementation, for example by specifying how to declare a free variable in Gurobi or what interface the program should provide to the team's existing code. Some instructions overlap between Math and Code entries, so Table~\ref{tab:stage-selection} compares the supplied groups rather than disjoint types of knowledge.

With reasoning enabled on these 64 models, the base LLM passed evaluation on 62 tasks without memory and 63 with system-provided memory, which also reduced the recorded token cost and mean elapsed time (Table~\ref{tab:memories}). Appendix~\ref{ec:membench} reports the full five-configuration comparison, together with supplementary GPT-6.1 Terra results (Table~\ref{tab:terra-memory}).

User-customized memory also communicates requirements of an existing code library that are not specified by the optimization model. For a two-supplier procurement model, we retained System-provided Math guidance and selected a personal Code entry specifying a builder name, a Gurobi model name, and a prefix for printed output. The generated program followed the function and model naming requirements but omitted the requested output prefix. Its optimization result remained correct. The program minimized $4+\|(x_1,x_2)\|_2$ subject to $x_1+x_2=2$ and $x\ge0$. It returned $(1,1)$ with cost $4+\sqrt2\approx5.414214$, the analytical optimum (Appendix~\ref{ec:edit-cases}).

\subsection{Evaluating the Pointwise Check}\label{sec:eval-check}

A five-variable polyhedral RO program generated by LLM on the 64-model benchmark returned the correct optimum even though it admitted infeasible decisions. Its value matched the reference optimum of $-12.60$ and agreed with the original model at both the returned decision and the reference optimizer. At 54 of 205 tested decisions, however, the program and the original model disagreed. For example, the program assigned a finite cost of $10.98$ to $\hat x=(-1.8,0,0,0,0)$. Even under zero shock this decision violates original constraint 5, since $7.6(-1.8)=-13.68<-12.54$. In constraint~5, the program added its expression for the worst-case penalty to the left-hand side of the lower-bound constraint. Increasing the auxiliary variables could therefore satisfy the row even when the original decision was infeasible.

As a controlled comparison, we modified each of the 256 benchmark models once to represent a possible reformulation error. The changes replaced a robust constraint by a nominal one, transposed an ellipsoid factor, relaxed moment conditions, or changed a Wasserstein support or transport norm. Of these, 215 differed from the original model in feasibility or cost at some decision, and 211 of those had finite optimal values (Table~\ref{tab:mutations}). Comparing optimal values (D0) detected 136. Comparing values or feasibility at the returned decision (D1) detected 103, and testing multiple decisions (D2) detected 181. Among the 74 changes missed by both D0 and D1, D2 detected 55.

Testing multiple decisions and comparing optimal values revealed different errors. The optimal-value comparison detected 11 changes missed by D2, bringing their combined count to 192 of 211. This complementarity was particularly clear for ellipsoidal changes, where D2 found 23 of 49, D0 found 28, and together they found 33. Disagreements near a curved feasibility boundary can be hard to sample. In a separate offline comparison across five seeds, boundary bisection detected 40--46 of the 49 ellipsoidal changes within 50 trial decisions, compared with 19--21 for mixed sampling. With 200 trials, bisection detected 48--49 (Appendix~\ref{ec:check-study}).

A check report can also guide a correction even when the decision itself stays the same. In an ellipsoidal portfolio task from the 64-model benchmark, the generated program incorrectly formed the transformed vector as \(w=Lx\) rather than \(w=L^\top x\). Its rotated-cone constraints then represented \(\|w\|_2^2/2\), rather than the required norm \(\|w\|_2\). It reported a guarantee of 13.654 at an allocation for which the original model guarantees 13.196. D1 supplied an adverse scenario and D2 flagged 121 of 205 decisions. After two repair rounds, the program used the transposed factor and the correct norm constraints. It returned the same allocation with the corrected guarantee of 13.196 and agreed at all 205 tested decisions.

Across the five LLM configurations on the 64-model set, the first check flagged 93 runs spanning 58 distinct tasks. Of those 93 runs, 26 passed evaluation after one or two repair rounds, including three with reasoning and No memory. Some corrections required revisiting the counterpart itself. In a polyhedral task a scenario violated constraint 5 by 1.17, and the system traced the violation to dual signs and stopped code generation. In a separate continuation we edited the support-function signs, budget-dual scaling, and one-sided-set constraints in the displayed counterpart. The resulting program attained the independently verified optimum $-7.634134$ (Appendix~\ref{ec:edit-cases}).

The memory runs used baseline parameters. Separate executable controls examined parameter changes (\Dt), which exposed four defects missed at the baseline. Appendix~\ref{ec:check-study} gives these controls, including incomplete comparisons and numerical flags.

The original model used for comparison also matters. In selected IndustryOR cases, the displayed models and their programs shared omissions relative to references independently constructed from the problem descriptions. These omissions escaped comparison with the displayed model but were exposed by the independent reference, including an order using more transport modes than allowed. This illustrates why reformulation checking complements review of the original model against the intended problem (Appendix~\ref{ec:published}).

%% file: figures/fig_memcost.tex
\begin{figure}[!hb]
\centering
\begin{tikzpicture}
\begin{axis}[
  name=left, width=0.47\linewidth, height=4.6cm,
  xlabel={Token cost for 256 tasks (USD)}, ylabel={Tasks passing evaluation},
  xmin=1.4, xmax=12.8, xtick={2,4,6,8,10,12}, ymin=50, ymax=280, ytick={64,128,192,256},
  tick label style={font=\footnotesize}, label style={font=\footnotesize},
  axis lines=left, clip=false,
]
\addplot[only marks, mark=o, mark size=3pt, black, thick] coordinates {(6.67149100,81)};
\addplot[only marks, mark=triangle, mark size=3.6pt, black, thick] coordinates {(4.33624421,188)};
\addplot[only marks, mark=square, mark size=2.8pt, black, thick] coordinates {(2.89903547,241)};
\addplot[only marks, mark=*, mark size=3pt, black] coordinates {(9.38032491,255)};
\node[font=\scriptsize, anchor=west] at (axis cs:7.0,81) {none};
\node[font=\scriptsize, anchor=west] at (axis cs:4.77,188) {expert rules};
\node[font=\scriptsize, anchor=south west] at (axis cs:3.13,244) {system-provided};
\node[font=\scriptsize, align=center, anchor=north] at (axis cs:9.38,246) {none,\\reasoning};
\end{axis}
\begin{axis}[
  at={(left.east)}, anchor=west, xshift=0.8cm,
  width=0.47\linewidth, height=4.6cm,
  xlabel={Mean minutes per task}, ylabel={},
  xmin=0, xmax=7.4, ymin=50, ymax=280, ytick={64,128,192,256},
  tick label style={font=\footnotesize}, label style={font=\footnotesize},
  axis lines=left, clip=false,
]
\addplot[only marks, mark=o, mark size=3pt, black, thick] coordinates {(1.91148,81)};
\addplot[only marks, mark=triangle, mark size=3.6pt, black, thick] coordinates {(1.38081,188)};
\addplot[only marks, mark=square, mark size=2.8pt, black, thick] coordinates {(0.69463,241)};
\addplot[only marks, mark=*, mark size=3pt, black] coordinates {(5.67365,255)};
\node[font=\scriptsize, anchor=west] at (axis cs:2.1,81) {none};
\node[font=\scriptsize, anchor=west] at (axis cs:1.6,188) {expert rules};
\node[font=\scriptsize, anchor=south west] at (axis cs:0.9,244) {system-provided};
\node[font=\scriptsize, align=center, anchor=north] at (axis cs:5.67,246) {none,\\reasoning};
\end{axis}
\end{tikzpicture}
\caption{Evaluation passes, token cost, and elapsed time on the same 256 tasks. Open markers indicate that reasoning was disabled. The filled marker denotes reasoning enabled.}
\label{fig:memcost}
\end{figure}
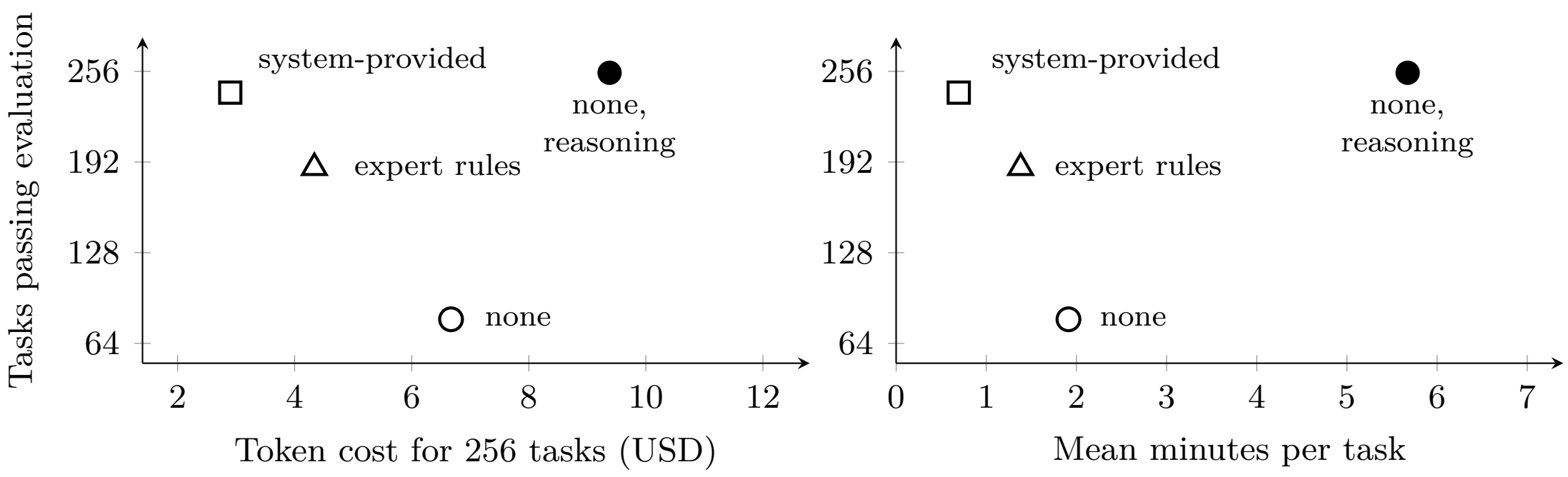

%% file: sections/07_conclusion.tex
\section{Conclusions}\label{sec:conclusion}

RobustOPT brings model confirmation, mathematical reformulation, solver implementation, and decision revision into one human--AI collaborative system for RO and DRO. Users determine the operational requirements and the treatment of uncertainty, while an LLM assists with formulation, derivation, and coding. The original model, explicit reformulation, and code remain available for inspection and editing as the problem develops.

The three components address different parts of this work. The interactive workflow connects the user's modeling choices with later revisions to data and fixed decisions. Independently selectable Math and Code memory let users apply mathematical knowledge while specifying the interfaces and solver conventions required by their own code libraries. The pointwise check compares the original model and its reformulation at common decisions, returning evidence that can guide a correction. Together, these components give users control over both the mathematical model and the instructions used to construct its implementation.

The application cases show how changes to uncertainty assumptions affect the solution and how users can reoptimize remaining orders while keeping selected quantities fixed. The experiments show that the memory increased LLM's evaluation passes with reasoning disabled. This configuration incurred lower token cost and took less time than reasoning alone, which achieved a higher pass rate. In a portfolio case, the pointwise check identified an overstated guaranteed return and guided a correction without changing the optimal allocation.

Future work could integrate review-guided continual learning into the memory module, using accumulated check results and user corrections to refine reusable Math and Code knowledge while retaining human oversight. Methodologically, extending the pointwise check to multistage adaptive problems would require tractable bounds on recourse costs after initial decisions are fixed, as well as broader evaluators for chance constraints and nonconvex losses. More broadly, we aim to develop RobustOPT as a platform for the operations research community by expanding its reviewed memory libraries across uncertainty classes and solver backends, and supporting longitudinal studies of how practitioners and students formulate, audit, and revise decisions under uncertainty.

%% file: tables/main_tables.tex
\input{tables/related}

\begin{table}[!htbp]
\centering\small
\caption{Mathematical and implementation guidance for an ellipsoid.}
\label{tab:entry}
\begin{tabularx}{\linewidth}{@{}P{1.55cm}LL@{}}
\toprule
 & Math entry & Code entry \\
\midrule
Purpose & Derive the worst-case expression. & Implement that expression in Gurobi. \\
Rule & For $\delta=Lu$ and $\|u\|_2\le1$, the largest deviation of $\delta^\top v$ is $\|L^\top v\|_2$. & Write $w_i=\sum_j L_{ji}v_j$ and $\sum_i w_i^2\le t^2$, with $t\ge0$. Set the bounds of free variables explicitly. \\
Scope & Ellipsoidal RO. & Ellipsoidal RO in Gurobi. \\
\bottomrule
\end{tabularx}
\end{table}

\input{tables/families}

\input{tables/evidence}

\input{tables/levels}

\input{tables/suite}

\input{tables/reuse}

\input{tables/memory-final4}

\input{tables/stage-selection}

\input{tables/mutations}
\FloatBarrier

%% file: tables/related.tex
\begin{table}[!htbp]
\centering
\caption{Comparison among LLM-based modeling systems. Before, During, and After refer to user interaction before reformulation, during reformulation, and after solving, respectively. Entry maintenance refers to adding, editing, or removing entries in a user's memory library. Math/Code selection denotes independent choices of Math memory and Code memory.}
\label{tab:related}
\footnotesize
\begin{tabular}{@{}P{2.15cm}P{2.05cm}C{1.0cm}C{1.0cm}C{0.8cm}C{1.9cm}C{1.7cm}C{1.55cm}C{1.7cm}@{}}
\toprule
 & & \multicolumn{3}{c}{User interaction} & \multicolumn{2}{c}{Memory customization} & \multicolumn{2}{c}{Reformulation check} \\
\cmidrule(lr){3-5}\cmidrule(lr){6-7}\cmidrule(l){8-9}
System & Problems & Before & During & After & Entry\newline maintenance & Math/Code\newline selection & Execution\newline check & Pointwise\newline check \\
\midrule
\mbox{OptiMUS-0.3} & Deterministic & $\checkmark$ & $\checkmark$ & & & & $\checkmark$ & \\
OptiGuide & Deterministic & & & $\checkmark$ & & & $\checkmark$ & \\
DAOpt & RO, DRO & & & & & & $\checkmark$ & \\
LLM4DRO & DRO & $\checkmark$ & & & & & $\checkmark$ & \\
\midrule
RobustOPT & RO, DRO & $\checkmark$ & $\checkmark$ & $\checkmark$ & $\checkmark$ & $\checkmark$ & $\checkmark$ & $\checkmark$ \\
\bottomrule
\end{tabular}

\end{table}

%% file: tables/families.tex
\begin{table}[!htbp]
\centering\small
\caption{Representative worst-case evaluations at a fixed decision.}
\label{tab:families}
\begin{tabularx}{\linewidth}{@{}P{2.45cm}P{5.0cm}L@{}}
\toprule
Family & Uncertainty or ambiguity set & Worst case at a fixed decision \\
\midrule
Box RO & $\underline\xi\le\xi\le\bar\xi$ & choose each endpoint by the sign of its coefficient \\
Budgeted RO & $\{\xi^0+d\circ\zeta:\|\zeta\|_\infty\le 1,\ \|\zeta\|_1\le\Gamma\}$ & sort the exposures $d_j|a_j(x)|$ and spend the budget on the largest \\
Polyhedral RO & $\{\xi:F\xi\le h\}$ or a finite scenario list & one linear program, or enumeration of the scenarios \\
Ellipsoidal RO & $\{\xi^0+\delta:\delta^\top Q^{-1}\delta\le\rho^2\}$ & closed form $a(x)^\top\xi^0+b(x)+\rho\|Q^{1/2}a(x)\|_2$ \\
Moment DRO & distributions on a support given by bounds, linear rows, a norm ball, or a finite point list, with affine moment conditions ($=$, $\le$, $\ge$) and upper bounds on $\E|\cdot|$ or $\E(\cdot)^2$ of affine functions & convex support: a finite convex program; finite support: an LP over the listed points \\
Wasserstein DRO & type-1 ball of radius $\varepsilon$ around an empirical distribution, $\ell_1$, $\ell_2$, or $\ell_\infty$ metric, support as above & convex support: a finite convex program; finite support: a transport LP \\
\bottomrule
\end{tabularx}

\smallskip
\parbox{\linewidth}{\footnotesize The examples use static decisions, RO losses of the form $a(x)^\top\xi+b(x)$, and DRO losses of the form $\max_g\min_l(a_{gl}(x)+b_{gl}(x)^\top\xi)$. Here $\xi^0$ is the nominal center, $d\ge0$, $\Gamma\ge0$, $Q\succ0$, and $\rho\ge0$. The uncertainty and ambiguity sets are assumed nonempty. The displayed DRO evaluations are LPs for polyhedral support, moment functions, and metrics, and second-order cone programs otherwise. Oracle coverage and attainment conditions are detailed in \ref{ec:oracles}.}
\end{table}

%% file: tables/evidence.tex
\begin{table}[!htbp]
\centering\small
\caption{Evidence for reformulation reuse and numerical checking.}
\label{tab:evidence}
\begin{tabularx}{\linewidth}{@{}P{6.0cm}L@{}}
\toprule
Evidence & Establishes \\
\midrule
Derivation of each rule under its premises & validity if the rules are also applied, combined, and implemented correctly \\
Every premise of the derivation holds at $\theta$ & the current data lie in the region covered by the derivation \\
Completed, unflagged comparisons at the tested decisions and parameters & agreement of the two numerical evaluations at those points within the stated tolerances \\
\bottomrule
\end{tabularx}
\end{table}

%% file: tables/levels.tex
\begin{table}[!htbp]
\centering\small
\caption{Checks and the evidence they provide.}
\label{tab:levels}
\begin{tabularx}{\linewidth}{@{}P{1.0cm}P{5.6cm}L@{}}
\toprule
Level & Question & What is compared \\
\midrule
M & Is the original model feasible? & nonempty sets; a feasible decision or an infeasibility certificate \\
S & Does the program run and solve? & execution and solver status \\
D1 & Is the returned decision as good as reported? & feasibility and worst-case cost at the returned decision \\
D2 & Does the program agree at other decisions? & feasibility and cost at trial decisions \\
\Dt & Does the agreement survive a change of data? & S, D1, and D2 at changed parameter values \\
\midrule
D0 & Is the optimal value right? (needs a reference) & the two optimal values \\
\bottomrule
\end{tabularx}
\end{table}

%% file: tables/suite.tex
\begin{table}[!htbp]
\centering\small
\caption{Workflow cases and outcomes. Complete-input runs and open-choice cases are separate; the latter include a continuation after editing.}
\label{tab:suite}
\begin{tabularx}{\linewidth}{@{}LrL@{}}
\toprule
Input or request & Units & Observed outcome \\
\midrule
Complete descriptions & 40 tasks & 40 first-attempt evaluation passes \\
Missing information & 5 controls & 5 clarification requests \\
Invalid information & 5 controls & 5 correction requests \\
Infeasible or unbounded model & 2 controls & 2 correct solver statuses \\
Open protection choices & 4 tasks & 4 clarifications; N03 completed after a subsequent model edit \\
Cumulative post-solve changes & 4 chains & 16 verified revisions (9 retained programs, 7 regenerated copies) \\
\bottomrule
\end{tabularx}
\end{table}

%% file: tables/reuse.tex
\begin{table}[!htbp]
\centering\small
\caption{Optimal worst-case costs after cumulative changes. R denotes a retained program and B a regenerated copy. Each column retains the changes in the preceding column. Each group is fixed at predeclared quantities from the original case baseline.}
\label{tab:reuse}
\resizebox{\linewidth}{!}{\begin{tabular}{@{}llrrrrrr@{}}
\toprule
Task & Application (unit) & $n$ & Baseline & New data & New protection & Fix group 1 & Fix groups 1, 2 \\
\midrule
N05 & procurement (million CNY) & 50 & 2,717.82 & 3,045.03 R & 3,282.13 R & 3,349.66 R & 3,427.30 R \\
N06 & power dispatch (CNY) & 50 & 8,404.04 & 9,386.22 R & 11,857.08 R & 12,077.14 R & 12,326.62 R \\
N07 & stock allocation (thousand CNY) & 50 & 799.29 & 865.50 B & 969.33 B & 983.51 B & 999.81 B \\
N08 & freight (thousand CNY) & 80 & 557.08 & 599.75 R & 670.90 B & 694.47 B & 731.95 B \\
\bottomrule
\end{tabular}}
\end{table}

%% file: tables/memory-final4.tex
\begin{table}[!htbp]
\centering\small
\caption{Memory and reasoning on 256 tasks starting from original models.}
\label{tab:memory-final4}
\setlength{\tabcolsep}{3pt}
\resizebox{\linewidth}{!}{\begin{tabular}{@{}llccccccrr@{}}
\toprule
& & \multicolumn{4}{c}{Evaluation passes by family (out of 64)} & \multicolumn{2}{c}{All tasks (out of 256)} & \multicolumn{2}{c}{Cost and time} \\
\cmidrule(lr){3-6}\cmidrule(lr){7-8}\cmidrule(l){9-10}
Guidance & Reasoning & Polyhedral & Ellipsoidal & Moment & Wasserstein & $\EP$ & $\OA$ & USD & Min./task \\
\midrule
No memory & off & 23 & 43 & 5 & 10 & 81 & 93 & 6.67 & 1.91 \\
Expert rules & off & 31 & 48 & 55 & 54 & 188 & 198 & 4.34 & 1.38 \\
System-provided & off & 53 & 64 & 62 & 62 & 241 & 242 & 2.90 & 0.69 \\
\midrule
No memory & on & 64 & 64 & 63 & 64 & 255 & 255 & 9.38 & 5.67 \\
\bottomrule
\end{tabular}}
\par\smallskip
\parbox{\linewidth}{\footnotesize Each configuration was run once per task, including permitted repairs. Cost is the estimated token cost for all 256 tasks. Time is mean elapsed minutes per task.}
\end{table}

%% file: tables/stage-selection.tex
\begin{table}[!htbp]
\centering\small
\caption{Math and Code memory on 64 tasks starting from original models. Reasoning is disabled. Each task has one attempt, including permitted repairs.}
\label{tab:stage-selection}
\begin{tabular}{@{}lrrrr@{}}
\toprule
Guidance used & Neither & Math only & Code only & Math + Code \\
\midrule
Evaluation passes ($\EP$/64) & 15 & 50 & 13 & 55 \\
Objective agreement ($\OA$/64) & 15 & 51 & 13 & 55 \\
Mean minutes per task & 2.10 & 1.15 & 2.36 & 1.20 \\
Token cost (USD) & 1.74 & 1.03 & 1.80 & 1.03 \\
\bottomrule
\end{tabular}
\end{table}

%% file: tables/mutations.tex
\begin{table}[!htbp]
\centering\small
\caption{Controlled model changes detected by optimal-value comparison (D0), the returned decision (D1), and multiple-decision testing (D2).}
\label{tab:mutations}
\begin{tabularx}{\linewidth}{@{}Lrrrrrr@{}}
\toprule
Change & Non-equivalent / total & \shortstack{Finite-optimum\\cases} & D0 & D1 & D2 & D0$\cup$D2 \\
\midrule
Polyhedral RO, robust constraint made nominal & 48/64 & 48 & 29 & 27 & 44 & 45 \\
Ellipsoidal RO, factor transposed & 49/64 & 49 & 28 & 16 & 23 & 33 \\
Moment DRO, mean equality made an upper bound & 39/39 & 39 & 33 & 30 & 39 & 39 \\
Moment DRO, upper mean bound removed & 19/19 & 19 & 10 & 4 & 19 & 19 \\
Moment DRO, variance bound removed & 6/6 & 2 & 2 & 0 & 2 & 2 \\
Wasserstein DRO, support widened & 36/46 & 36 & 16 & 8 & 36 & 36 \\
Wasserstein DRO, ground norm changed & 18/18 & 18 & 18 & 18 & 18 & 18 \\
\midrule
Total & 215/256 & 211 & 136 & 103 & 181 & 192 \\
\bottomrule
\end{tabularx}

\smallskip
\parbox{\linewidth}{\footnotesize The first count column gives the number of changes that alter feasibility or cost at some decision out of all changes made. Detection counts use only changed models with finite optima. D2 includes the changed model's optimizer, four neighbors, available box corners, and 40 further decisions, without the reference optimizer.}
\end{table}

%% file: appendices/companion.tex
\input{appendices/ec_proofs}
\input{appendices/ec_check}
\input{appendices/ec_memory}
\input{appendices/ec_ledger}

%% file: appendices/ec_proofs.tex
\section{Proofs, Reference Derivations, and Oracles (Supporting Section~\ref{sec:check})}\label{ec:proofs-all}

\subsection{Properties of Records and Pointwise Comparisons}\label{ec:proofs}
\def\theHproposition{EC.\arabic{proposition}}

We give the detailed statements and proofs of the two properties used in Section~\ref{sec:check}, then specify when checking additional decisions subsumes an optimal-value comparison. Throughout, $\theta$ is fixed and omitted where no confusion arises. Feasible points of a finite program have finite objective values. Recall that $g(x)=\inf\{G_0(z):z\in\cG,\ \varphi(z)=x\}$ and $\Fbar(x)=F_0(x)$ for $x\in\cX$ and $+\infty$ otherwise, both defined for every $x\in\R^n$.

\begin{repeatproposition}[Proposition~\ref{prop:record} (detailed statement).]
Consider a record $\cR=(P,\Pi,C,\varphi)$ and $\theta\in\ThetaPi$.
\begin{enumerate}[label=(\alph*),leftmargin=*]
\item \emph{Decision equivalence.} If $C(\theta)$ is exact for $P(\theta)$, the two have the same optimal value; $\varphi$ maps every optimal solution of $C(\theta)$ to an optimal decision of $P(\theta)$; and every optimal decision $x^\star$ of $P(\theta)$ at which the infimum in \eqref{eq:projected} is attained is the image of an optimal solution of $C(\theta)$.
\item \emph{Reuse and decision fixing.} If the record is valid, part (a) holds at every $\theta'\in\ThetaPi$ for the retained program $C(\theta')$. For every restriction of the original decisions alone, such as $x_j=\bar x_j$ for $j\in J$, the restricted program is exact for the restricted original model.
\item \emph{Conservative programs.} If $C(\theta)$ is only conservative for $P(\theta)$, every decision it accepts is feasible for $P(\theta)$ and has a true worst-case cost no larger than the reported one, also under the restrictions in (b); the returned decision need not be optimal.
\end{enumerate}
\end{repeatproposition}

\proof{Proof of Proposition~\ref{prop:record}.}
(a) Exactness gives $g=\Fbar$ on $\R^n$. Since every $z\in\cG$ has $\varphi(z)\in\R^n$, $\val C=\inf_{z\in\cG}G_0(z)=\inf_{x\in\R^n}g(x)=\inf_{x\in\R^n}\Fbar(x)=\val P$. Let $z^*$ be optimal for $C$ and $x^*=\varphi(z^*)$. Then $\Fbar(x^*)=g(x^*)\le G_0(z^*)=\val C=\val P<\infty$, so $x^*\in\cX$ and $F_0(x^*)\le\val P$; thus $x^*$ is optimal for $P$. Conversely, let $x^\star$ be optimal for $P$ and let $z^\star$ attain $g(x^\star)$. Then $z^\star\in\cG$ and $G_0(z^\star)=g(x^\star)=F_0(x^\star)=\val P=\val C$, so $z^\star$ is optimal for $C$.

(b) For $\theta'\in\ThetaPi$, validity of the record states that $C(\theta')$ is exact for $P(\theta')$, and part (a) applies. Let $Y\subseteq\R^n$ be defined by constraints on $x$ alone. The restricted program has projected value $g_Y(x)=g(x)$ for $x\in Y$ and $+\infty$ otherwise, because the added constraint does not involve the other program variables. The restricted original problem has feasible set $\cX\cap Y$ and objective $F_0$, so its extended value is $\Fbar$ on $Y$ and $+\infty$ otherwise. Hence the two agree on $\R^n$, which is exactness of the restricted program.

(c) If $g\ge\Fbar$ on $\R^n$ and $z$ is feasible for $C$ with $x=\varphi(z)$, then $\Fbar(x)\le g(x)\le G_0(z)<\infty$, so $x\in\cX$ and $F_0(x)\le G_0(z)$. Under a restriction to $Y$, $g_Y=g$ on $Y$ as in (b), so the same holds. That the returned decision need not be optimal is shown by Example~\ref{ex:shared} at $a=0$: $g\equiv 1$ on $[0,1]$, so $x=0$ is optimal for $C$, whereas $F_0(0)=1>1/2=\val P$.\Halmos
\endproof

\parhead{Claims of Example~\ref{ex:shared}.} The buyer's cost $ax+x\xi+(1-x)(1-\xi)=ax+(1-x)+(2x-1)\xi$ is affine in $\xi$, so its supremum over distributions on $[0,1]$ is attained by a point mass at an endpoint and equals $ax+\max\{x,1-x\}$. The separated program bounds the two surcharges by their own suprema, $x$ and $1-x$, whose sum is $1$. Thus $g(x;a)=ax+1\ge F_0(x;a)$ on $[0,1]$, so the separated program is conservative; it is not exact because $g(1/2;a)-F_0(1/2;a)=1/2$. At $a=2$, $F_0(x;2)=1+x$ on $[0,1/2]$ and $3x\ge 3/2$ on $[1/2,1]$, so $\val P=1$ at $x=0$, and $g(x;2)=2x+1$ also has minimum $1$ at $x=0$. At $a=0$, $\val P=1/2$ at $x=1/2$, whereas $g\equiv 1$. The two cases differ only in the deterministic premium $a$.

\begin{repeatproposition}[Proposition~\ref{prop:counter} (detailed statement).]
Let $\hat x\in\R^n$.
\begin{enumerate}[label=(\alph*),leftmargin=*]
\item Suppose some $\hat z$ with $\varphi(\hat z)=\hat x$ is feasible for $C(\theta)$. If $\hat x\notin\cD(\theta)$, or a scenario $\hat\xi\in\cU(\theta)$ (or a distribution $\hat\Prob\in\cP(\theta)$) violates a constraint of $P(\theta)$ at $\hat x$ or yields a cost above $G_0(\hat z;\theta)$, then $C(\theta)$ is neither exact nor conservative.
\item If $\hat x\in\cX(\theta)$ and $F_0(\hat x;\theta)\le u_0$, and a lower bound $\ell\le g(\hat x;\theta)$ satisfies $\ell>u_0$, then $C(\theta)$ is not exact.
\end{enumerate}
\end{repeatproposition}

\proof{Proof of Proposition~\ref{prop:counter}.}
(a) Since $\hat z\in\cG$ and $\varphi(\hat z)=\hat x$, $g(\hat x)\le G_0(\hat z)<\infty$. If $\hat x\notin\cD$, then $\Fbar(\hat x)=+\infty>g(\hat x)$. If $f_i(\hat x,\hat\xi)>0$ for some $i\ge 1$ and $\hat\xi\in\cU$, then $F_i(\hat x)>0$, so $\hat x\notin\cX$ and again $\Fbar(\hat x)=+\infty>g(\hat x)$. If instead $G_0(\hat z)<f_0(\hat x,\hat\xi)\le F_0(\hat x)$, then $g(\hat x)<\Fbar(\hat x)$. Each case violates both exactness and conservativeness. In DRO, $\hat\Prob\in\cP$ gives $\E_{\hat\Prob}f_i(\hat x,\xi)\le F_i(\hat x)$ for every $i$, and the same argument applies. (b) Exactness would require $g(\hat x)=\Fbar(\hat x)=F_0(\hat x)\le u_0<\ell\le g(\hat x)$, a contradiction.\Halmos
\endproof

\begin{proposition}[Conditional inclusion of comparisons]\label{prop:hier}
Suppose both problems have finite optimal values, the program is solved exactly with returned decision $x^*$, and $P(\theta)$ has an optimal decision $x^{\mathrm{ref}}$. Use exact pointwise values and feasibility, with tolerance $\epsilon(1+|\Fbar(x;\theta)|)$ for finite values and $\epsilon(1+|\val P(\theta)|)$ for D0, where $0\le\epsilon<1$. If the decisions examined by D2 include $x^*$ and $x^{\mathrm{ref}}$, then D2 flags every program that D0 or D1 flags. If \Dt{} examines the baseline parameters with the decisions of D2, it flags every program that D2 flags.
\end{proposition}

\proof{Proof of Proposition~\ref{prop:hier}.}
D1 compares $\Fbar(x^*)$ with the reported value $g(x^*)=\val C$, which is D2's comparison at $x^*$. If D0 flags with $\val C<\val P$, its gap satisfies $\val P-\val C>\epsilon(1+|\val P|)$, while optimality of $P$ gives $\Fbar(x^*)\ge\val P$. If $x^*$ is infeasible or $\Fbar(x^*)=+\infty$, D2 flags it directly. Otherwise, define the gap function $H(u)=u-\val C-\epsilon(1+|u|)$. It is strictly increasing because $\epsilon<1$. Hence $H(\Fbar(x^*))\ge H(\val P)>0$, so D2 flags the finite-value discrepancy at $x^*$. If D0 flags with $\val C>\val P$, then $g(x^{\mathrm{ref}})\ge\val C>\val P=\Fbar(x^{\mathrm{ref}})$ beyond the same tolerance, because $\val C=\inf_x g(x)$; D2 flags at $x^{\mathrm{ref}}$. If \Dt{} examines the baseline parameters with the test set of D2, it repeats every comparison of D2.\Halmos
\endproof

\subsection{Reference Derivations for the Four Families}\label{ec:derivations}

\parhead{Ellipsoidal RO (walkthrough).} Let $\cU=\{\delta:\delta^\top Q^{-1}\delta\le 1\}$ with $Q\succ 0$, and write $\delta=Q^{1/2}u$ with $\|u\|_2\le 1$. For any $x$, $\min_{\delta\in\cU}\delta^\top x=\min_{\|u\|_2\le 1}u^\top Q^{1/2}x=-\|Q^{1/2}x\|_2=-\sqrt{x^\top Qx}$, attained at $\delta^*(x)=-Qx/\sqrt{x^\top Qx}$ for $x\ne 0$. With $Q=LL^\top$, $\sqrt{x^\top Qx}=\|L^\top x\|_2$, so the worst-case return is $c^\top x-\|L^\top x\|_2$, and the variables $w=L^\top x$, $\|w\|_2\le t$, $t\ge 0$ give the program of Section~\ref{sec:walkthrough}. After negating the maximization objective, its projected value equals the negative worst-case return at every feasible $x$, and the program accepts exactly the decisions of the original model, so it is exact whenever $Q\succ 0$. The transpose matters: $\|Lx\|_2\ne\|L^\top x\|_2$ in general. For the walkthrough data, $L=\begin{psmallmatrix}0.03&0\\0.00667&0.03944\end{psmallmatrix}$; at $x=(0,1)$, $\|Lx\|_2=0.0394$ while $\|L^\top x\|_2=0.04$, the incorrect implementation using \(Lx\) therefore reports a guaranteed return of 4.06\%, whereas the original model guarantees 4.00\%, and $\delta^*=(-0.005,-0.040)$ is the witness. Omitting $t\ge 0$ leaves $w^\top w\le t^2$, which admits $t\le-\|w\|_2$ and makes the maximization unbounded.

\parhead{Box and budgeted RO.} Consider a row $\tilde a^\top x\le b$ with $\tilde a_j=a_j+d_j\zeta_j$, $|\zeta_j|\le 1$, and $\sum_j|\zeta_j|\le\Gamma$. Its worst case is $a^\top x+\max\{\sum_j d_j|x_j|\zeta_j:0\le\zeta\le 1,\ \sum_j\zeta_j\le\Gamma\}$, and LP duality gives
\[
a^\top x+\min_{\pi\ge 0,\,p\ge 0}\Big\{\Gamma\pi+\sum_j p_j:\ \pi+p_j\ge d_jy_j\ \forall j\Big\},\qquad y_j\ge x_j,\ y_j\ge-x_j,
\]
which is exact for every $\Gamma\ge 0$. Here $\pi$ and $p_j$ are the multipliers for the budget and coordinate bounds, respectively, and $y$ bounds $|x|$ componentwise. One may choose $y=|x|$, or simply $y=x$ when $x\ge0$. When $\Gamma$ is at least the number of uncertain coefficients, the budgeted set is the box and $\pi=0$ is optimal. The endpoint rule for a box replaces $\tilde a_j$ by $a_j+d_j$; it is exact only for $x_j\ge 0$, so its record carries the premise that the lower bounds of $x$ are nonnegative. For $a\in[1.5,2.5]\times[0.75,1.25]$ and $x=(-1,2)$, the worst case is $1.5\cdot(-1)+1.25\cdot 2=1$, whereas the upper endpoints give $2.5\cdot(-1)+1.25\cdot 2=0$.

\parhead{Moment DRO on an interval.} Let $\xi$ be scalar, $\ell(x,\xi)=\max_{k\le K}(a_k(x)\xi+b_k(x))$ convex piecewise affine in $\xi$, and $\cP=\{\Prob:\operatorname{supp}\Prob\subseteq[\underline s,\bar s],\ \E\xi=\mu,\ \E|\xi-\mu|\le b\}$. Conic linear duality gives
\[
\begin{aligned}
\sup_{\Prob\in\cP}\E_\Prob\ell(x,\xi)&=\min_{\lambda_0,\lambda_1\in\R,\ \lambda_2\ge 0}\Big\{\lambda_0+\lambda_1\mu+\lambda_2 b:\\
&\qquad \lambda_0+\lambda_1\xi+\lambda_2|\xi-\mu|\ge a_k(x)\xi+b_k(x)\ \ \forall\xi\in[\underline s,\bar s],\ k\le K\Big\}
\end{aligned}
\]
when $\underline s<\mu<\bar s$ and $b>0$ (Slater conditions). The multipliers $\lambda_0$, $\lambda_1$, and $\lambda_2$ correspond to normalization, the mean, and the mean absolute deviation bound. For each $k$, the difference of the two sides is piecewise affine in $\xi$ with the single breakpoint $\mu$, so the semi-infinite constraint holds if and only if it holds at $\xi\in\{\underline s,\mu,\bar s\}$. This yields an LP with $3K$ constraints whose atoms do not depend on $x$; the breakpoint of the loss, which may depend on $x$, need not be an atom because the constraints are imposed per affine piece. The dual of this LP is an LP over distributions on $\{\underline s,\mu,\bar s\}$, which is how the oracle computes the worst case and its witness. For a vector with box support and coordinatewise mean and mean absolute deviation constraints, each affine-piece inequality separates by coordinate. The corresponding primal LP uses the joint Cartesian grid of these three-point sets; it does not assume independent coordinates.

The mean must be among the atoms. For $\ell(x,\xi)=|x-\xi|$ on $[0,10]$ with $\mu=4$ and $b=2$, the worst case at $x=5$ is $8/3$, attained by $\Prob(0)=1/4$, $\Prob(4)=7/12$, $\Prob(10)=1/6$: its mean is $4$, its mean absolute deviation is $1+1=2$, and $\E|5-\xi|=5/4+7/12+5/6=8/3$. A faulty program omits the mean atom $\mu=4$ and restricts the distribution to $\{0,5,10\}$, the endpoints and the loss breakpoint. The mean and normalization conditions, $5p_5+10p_{10}=4$ and $p_0+p_5+p_{10}=1$, give $p_{10}=0.4-0.5p_5$ and $p_0=0.6-0.5p_5$. The deviation bound $4p_0+p_5+6p_{10}\le2$ then reduces to $4.8-4p_5\le2$, so $p_5\ge0.7$. Its worst-case loss is therefore $5(p_0+p_{10})=5(1-p_5)\le3/2$. This understates the true worst-case cost of $8/3$, and the distribution above supplies a counterexample under Proposition~\ref{prop:counter}(a). At $\mu=5$, by contrast, $\{0,5,10\}=\{\underline s,\mu,\bar s\}$ and the two programs coincide for every $b\ge0$, with value $m+(1-m/5)|x-5|$ for $x\in[0,10]$ and $m=\min\{b,5\}$: an error invisible at one parameter value and visible at another, which is what \Dt{} targets.

\parhead{Attainment at a zero-variance boundary.}\label{ec:variance-boundary} A dual used to replace a worst-expectation constraint must attain its value at the constraint boundary. For $\cP_0=\{\Prob:\E\xi=0,\ \E\xi^2\le0\}$, only $\xi=0$ almost surely is possible, so $\sup_{\Prob\in\cP_0}\E|\xi|=0$. The unreduced dual uses a free intercept $\alpha$, a free mean multiplier $y$, and a nonnegative second-moment multiplier $\rho$. It requires $\alpha+y\xi+\rho\xi^2\ge\pm\xi$ for every $\xi\in\R$. Introducing nonnegative auxiliary variables $r_+,r_-$ for the two affine pieces and bounding the dual value by $t$ gives
\[
\alpha\le t,\quad \alpha\ge r_+,r_-,\quad
(1-y)^2\le4\rho r_+,\quad(-1-y)^2\le4\rho r_-,\quad \rho,r_+,r_-\ge0.
\]
At $t=0$, these conditions force $r_+=r_-=0$ and simultaneously $y=1$ and $y=-1$. Every $t>0$ is feasible with $y=0$, $r_+=r_-=\alpha=t$, and $\rho=1/(4t)$. Thus the infimum is correct but the projected feasible set excludes $t=0$. The full-space variance rule applies directly when the variance bounds are strictly positive. For a zero-variance coordinate, the modeler can substitute the mean before applying the rule to the remaining coordinates. This reduction concerns dual attainment. The primal fixed-decision evaluation is given in \ref{ec:oracles}.

A fixed mean on a curved boundary can cause the same difficulty. On the unit disk, $\E\xi=(1,0)$ forces $\xi=(1,0)$ almost surely, so $\sup\E|\xi_2|=0$. Let $y=(y_1,y_2)$ be the multiplier on the mean constraint and $e_2=(0,1)$. Taking the support function over the unit disk, the unreduced mean dual minimizes $y_1+\max\{\|e_2-y\|_2,\|-e_2-y\|_2\}$. Its objective is at least $y_1+\sqrt{y_1^2+y_2^2+1}>0$ for every finite $y$, but tends to zero along $y=(-t,0)$ as $t\to\infty$. Substituting the forced point gives the attained value directly.

\parhead{Wasserstein DRO of type 1.} Let $\hat\Prob_N=\frac1N\sum_i\delta_{\hat\xi_i}$, a norm $\|\cdot\|$ with dual norm $\|\cdot\|_*$, a polyhedral support $\Xi=\{\xi:C\xi\le d\}$ containing the samples, and a loss $\max_k(a_k(x)^\top\xi+b_k(x))$. For $\varepsilon\ge 0$ \citep{mohajerin2018data},
\[
\begin{aligned}
\sup_{\substack{\Prob:W_1(\Prob,\hat\Prob_N)\le\varepsilon\\\operatorname{supp}\Prob\subseteq\Xi}}\E_\Prob\max_k\big(a_k^\top\xi+b_k\big)=\min_{\lambda,s,\gamma}\Big\{\lambda\varepsilon+\frac1N\sum_i s_i:\ & s_i\ge b_k+a_k^\top\hat\xi_i+\gamma_{ik}^\top(d-C\hat\xi_i),\\
& \|C^\top\gamma_{ik}-a_k\|_*\le\lambda,\ \gamma_{ik}\ge 0\ \ \forall i,k\Big\}.
\end{aligned}
\]
Here $\lambda$ is the multiplier on the transport budget, $s_i$ is the epigraph variable for sample $i$, and $\gamma_{ik}$ contains the support multipliers for sample $i$ and loss piece $k$. Without the support, $\gamma_{ik}=0$ and the condition becomes $\|a_k\|_*\le\lambda$. In a scalar inventory problem with order quantity $x$, demand support $\xi\ge 0$, overage cost $c_o$, and shortage cost $c_s$, the loss $\max\{c_o(x-\xi),c_s(\xi-x)\}$ has slopes $-c_o$ and $c_s$ in $\xi$. Each unit of transport budget raises the loss at rate $c_s$ when mass moves upward and at rate $c_o$ when it moves downward, and only downward transport is limited by the support. When $c_s>c_o$, upward transport alone can approach the full-space supremum, so the support is inactive and dropping it leaves the value unchanged at every decision; when $c_o>c_s$, downward transport can become worst, the support then bounds it, and dropping it can overstate the worst case. For a single sample at zero, order quantity zero, and radius one, the loss $\max\{-c_o\xi,c_s\xi\}$ has worst value $c_s$ on the nonnegative support and $\max\{c_s,c_o\}$ on the whole line. Removing the support leaves the value at $8$ for $(c_s,c_o)=(8,1)$ but changes it from $1$ to $8$ for $(1,8)$.

\parhead{Lifted representations in modeling packages.} A package can represent the same original model through a different but exact program, for example by lifting a piecewise affine loss into auxiliary random variables under an event-wise ambiguity set \citep{chen2020rsome}. Such a program serves as an independent reference only if its projected value agrees with the compact reformulation at every original decision. Eliminating the lifted variables reduces RSOME's program for the moment and Wasserstein examples above to the compact LPs. On 44 moment DRO models with polyhedral, budgeted, ball, ellipsoidal, and full-space supports, RSOME's native formulation and the routines of \ref{ec:oracles} agreed in all 132 evaluations, with a largest difference of $6.97\times10^{-7}$ relative to $1+|v|$.

\subsection{Worst-Case Routines}\label{ec:oracles}

The routines read the structured representation of the original model. Each coefficient is an expression over the decisions, the uncertain parameters, and the data. These expressions are parsed and evaluated numerically without executing generated code or calling an LLM.

\parhead{RO.} For a constraint or objective affine in $\xi$, the worst case is computed in closed form for box, budgeted, and ellipsoidal sets (Table~\ref{tab:families}), by one LP for a polyhedron, and by enumeration for a finite scenario list. The maximizer is the witness. A constraint required to hold for every realization of a DRO model is evaluated in the same way, over the support.

\parhead{DRO.} At a fixed decision $\hat x$, a piecewise linear loss is read from the expression tree as $h(\xi)=\max_g\min_l(a_{gl}+b_{gl}^\top\xi)$, with at most 4096 pieces. Moment functions $\psi_k$ may be affine (with $=$, $\le$, or $\ge$), or $|\cdot|$ or $(\cdot)^2$ of an affine function (with $\le$). The routine depends on the support $S$.

\emph{Convex support.} Let $S$ be closed and convex, given by bounds, linear rows $P\xi\le q$, a norm ball, or their intersection. For a moment set the worst case is
\[
\max\Big\{\sum_g t_g:\ p_g\ge0,\ \sum_g p_g=1,\ z_g\in p_gS,\ t_g\le p_ga_{gl}+b_{gl}^\top z_g\ \forall g,l,\ \text{moment conditions in }(p_g,z_g)\Big\},
\]
Here $p_g$ is the probability mass assigned to group $g$, $z_g$ is its unnormalized first moment, and $t_g$ is a lower bound on the group's contribution to expected loss. The constraint $z_g\in p_gS$ uses the closed perspective of $S$: for $p_g>0$ it means $z_g/p_g\in S$, and for $p_g=0$ it means that $z_g$ lies in the recession cone of $S$. Convex moment conditions use perspective constraints. For example, $\E(a+b^\top\xi)^2\le c$ becomes $\sum_g w_g\le c$ with $(b^\top z_g+p_ga)^2\le w_gp_g$ and $w_g\ge0$. The nonnegativity is necessary also at $p_g=0$.

Collapsing the part of a distribution on which group $g$ attains $h$ to its conditional mean $z_g/p_g$ keeps every affine moment, does not increase a convex moment (Jensen's inequality), and does not lower $\E h$, because $\min_l$ of affine functions is concave; the conditional mean lies in $S$ because $S$ is convex. At a feasible solution with positive masses, the atoms $z_g/p_g$ yield an admissible distribution with expected loss at least $\sum_g t_g$. To approximate zero-mass recession terms, choose any admissible distribution with finite first moment and mean $\bar\xi$. If there are $G$ groups, the allocation $p_g^0=1/G$ and $z_g^0=p_g^0\bar\xi$ is feasible with suitable loss and moment auxiliaries, by Jensen's inequality. Mix the full feasible solution with this allocation using weight $0<\delta<1$. The resulting masses are positive, convexity preserves feasibility, and the objective converges to its original value as $\delta\downarrow0$. This establishes equality of the suprema. If every nonzero $z_g$ in an optimal solution has positive mass, the resulting atoms attain the worst case.

For a type-1 Wasserstein ball of radius $\varepsilon$ around $\sum_i w_i\delta_{\hat\xi_i}$ with convex support $S$ and an $\ell_1$, $\ell_2$, or $\ell_\infty$ metric, the same collapse per sample, which does not increase the convex transport cost, gives
\[
\begin{aligned}\max\Big\{\sum_i w_i\sum_g t_{ig}:\ &\alpha_{ig}\ge0,\ \textstyle\sum_g\alpha_{ig}=1,\ \alpha_{ig}\hat\xi_i-q_{ig}\in\alpha_{ig}S,\\ &t_{ig}\le\alpha_{ig}(a_{gl}+b_{gl}^\top\hat\xi_i)-b_{gl}^\top q_{ig},\ \textstyle\sum_i w_i\sum_g\|q_{ig}\|\le\varepsilon\Big\},\end{aligned}
\]
Here $\alpha_{ig}$ is the share of sample $i$ assigned to group $g$. For $\alpha_{ig}>0$, $q_{ig}/\alpha_{ig}$ is the displacement from the new atom to $\hat\xi_i$. The atoms $\hat\xi_i-q_{ig}/\alpha_{ig}$, for $\alpha_{ig}>0$, carry mass $w_i\alpha_{ig}$ \citep[cf.][Theorem 4.4]{mohajerin2018data}. Both programs are LPs when $S$, the moment functions, and the metric are polyhedral (solved by HiGHS) and conic programs otherwise (solved by ECOS).

\emph{Finite support.} The collapse requires convexity: for $S=\{-1,1\}$ and $\E\xi=0$, the distribution $\frac12\delta_{-1}+\frac12\delta_1$ is admissible, but its conditional mean $0$ is not in $S$. For a finite point list the routine therefore does not use the programs above. For a moment set it solves an LP over the probabilities of the listed points, with the moment conditions as linear constraints; for a Wasserstein ball it solves the transport LP from the samples to the listed points. Both are exact, and their solutions are finite worst-case distributions.

\emph{One-dimensional supports.} For a scalar uncertain parameter on a bounded interval with piecewise affine moment functions, the routine solves an LP over the endpoints and the breakpoints of the loss and moment functions. A quadratic moment upper bound instead uses the convex-support conic program above, even in one dimension. For scalar Wasserstein balls, the transport LP also includes the samples and, on an unbounded interval, variables for transport along an unbounded ray. Structures outside the supported classes are reported as unavailable, never as passed.

\parhead{Witnesses when the program's solution has no finite counterpart.} If an optimal solution of the convex-support program has $\alpha_{ig}=0$ with $q_{ig}\ne0$ (or $p_g=0$ with $z_g\ne0$), that solution does not correspond to a finite distribution: it describes mass sent to infinity along a ray. The value of the program is still the supremum, but a finite witness must satisfy the original distributional restrictions. For Wasserstein DRO, a single far point on the ray would exceed the transport budget. A finite witness for an understatement can instead be a distribution that moves a vanishing share $1/t$ of the sample's mass a distance proportional to $t$ along the ray, which stays within the budget and whose expected loss approaches the value as $t$ grows; for example, with $\hat\Prob=\delta_0$, $\Xi=\R_+$, $\varepsilon=1$, and $h(\xi)=(\xi-1)_+$, the distribution $(1-1/t)\delta_0+(1/t)\delta_t$ has distance $1$ and expected loss $1-1/t$ for $t>1$. For moment DRO, a finite witness must be checked against all moment conditions. An overstatement in this case is established by bounds, as in Proposition~\ref{prop:counter}(b), with the dual value of the routine's program as the upper bound.

\parhead{Validation.} At the reference decisions of the 256 benchmark models in Section~\ref{sec:eval-memory}, the routines reproduced the certified optimal values within an absolute error of $1.18\times10^{-7}$. Among 12,288 further decisions, 7,565 had a finite worst case; every returned scenario, distribution, or transport plan satisfied the stated restrictions and attained the returned value. At three further decisions per model, we recomputed the uncertain objectives and rows directly from the raw coefficients by independent methods (an LP in the uncertain coefficients for polyhedral RO, the quadratic form of the shape factor for ellipsoidal RO, a joint-grid LP or RSOME's native formulation for moment DRO, and a closed form or the support-aware dual for Wasserstein DRO); all 1,728 values agreed within $10^{-5}(1+|v|)$. The median time per decision, including the verification of the witness, was 1.10 ms for polyhedral RO, 0.45 ms for ellipsoidal RO, 0.60 ms for moment DRO, and 1.50 ms for Wasserstein DRO. On the 64 tasks of the memory benchmark, a cutting-plane method that uses only these routines to generate cuts reproduced the optimal values certified by RSOME, with every gap below $10^{-6}(1+|v_{\mathrm{LB}}|)$, where $v_{\mathrm{LB}}$ is the cutting-plane lower bound (\ref{ec:membench}).

\parhead{Coverage in the experiments.} The workflow evaluation distinguishes constraints on nominal values, constraints that hold throughout the support, and constraints on worst-case expectations. It represents a one-dimensional norm ball as an interval. The historical program study uses the check results obtained for those programs. Section~\ref{ec:ledger} describes the model representations and data examined in each study.

%% file: appendices/ec_check.tex
\section{Implementation of the Workflow and the Check (Supporting Sections~\ref{sec:workflow} and~\ref{sec:check})}\label{ec:checker-all}

\subsection{Model Representation and Decision Map}\label{ec:schema-map}

\parhead{Structured representation of the original model.} The routines of \ref{ec:oracles} read a machine-readable representation of the original model, including the decisions with their domains, the uncertain quantity with its set or ambiguity set, the objective, and the constraints, each marked as nominal, holding for every realization, or holding in expectation. Coefficients are expressions over decision symbols, the uncertain symbol, and the data, so that a change of data flows through without rewriting the model. The LLM drafts this representation from the original mathematical model after user confirmation. The check verifies that the representation refers to the current model and that every datum it uses is bound to a unique input. The routines then evaluate it without an LLM. The decision maker confirms the business facts, and the modeler checks that the structured representation preserves the displayed equations. An omission shared by this representation and the code cannot be found by comparing them.

\parhead{Numerical implementation.} In the interactive workflow, Step~4 uses an LLM-generated implementation prepared with the system's default solver and Code memory for its numerical comparisons. At Step~5, users can inspect the program, edit it, or request an implementation with their chosen solver and Code memory. The check report applies to the program and solver evaluated. After a change, a fresh check can be requested at Results. In the 40-task complete-input batches, delivery required a completed, unflagged check for the final program (Section~\ref{ec:suite}). Preparing the representations and revising code use the LLM; the worst-case evaluations and fixed-decision solves do not.

\parhead{Decision map and fixing.} The program declares each original decision as a solver variable, and the decision map records which variable represents which decision. To evaluate the program at $\hat x$, the check sets both bounds of the mapped variables to the requested values in a copy of the model, keeping the program's own bounds and variable types. A request outside a bound by more than $10^{-6}$ is rejected. For a smaller violation the program uses the bound itself (Section~\ref{ec:threshold-diagnostic}). After the solve, the original bounds are restored. If the map omits a decision, the comparison is incomplete. Certified infeasibility at the same $\hat x$ accepted by the original model with finite worst-case cost is a counterexample with $\ell=+\infty$ in Proposition~\ref{prop:counter}(b). An unbounded fixed program with a valid improving ray witnesses an understatement when the original value is finite.

\subsection{Checker Procedure and Reports}\label{ec:checker}

For each task, the report identifies the applicable levels in Table~\ref{tab:levels} and the comparisons that completed. Original-model evaluations require the loss, support, moment conditions, and row semantics accepted by the routines in \ref{ec:oracles}. Execution status S may be available even when an original-model comparison is not. D0 requires a supplied reference optimum, and \Dt{} is requested separately from the baseline check.

\parhead{Trial decisions.} D2 tests the returned decision $x^*$ and four neighbors $x^*+\epsilon\odot(|x^*|+1)$, where the coordinates of $\epsilon$ are independent $N(0,0.05^2)$ draws. It also tests all vertices of the variable box when the bounds are finite and $n\le4$. The random budget ranges from 20 to 300 decisions, with a default of 200. In continuous models with linear deterministic constraints, up to half of the budget is spent on extreme points and convex combinations of points of the deterministic polyhedron. The balance is drawn from the variable box, with integer coordinates rounded and a window of scale $|x^*_j|+1$ (or $1$ when $x^*$ is unavailable) for unbounded coordinates. These rounded trials test valid discrete assignments; they do not by themselves detect a program that incorrectly permits fractional decisions. If a share $p$ of a trial distribution lies in a region of disagreement beyond tolerance, $k$ independent draws from that distribution miss the region with probability $(1-p)^k$. This calculation concerns the independent draws only. The protocol also includes structured decisions.

\parhead{Trial parameters.} \Dt{} tries 20 parameter vectors formed from the data that both the program and the structured representation read, scaling the selected values by independent factors drawn uniformly from $[0.7,1.3]$ and renormalizing empirical probabilities. Vectors that violate a recorded premise are discarded. Structural threshold values, such as a zero radius, are tried separately. Each vector receives a fresh solve, level M, and D2 with 12 random decisions plus the structured points.

\parhead{Comparison.} At each decision, feasibility is compared first, then finite values within $10^{-5}(1+|\Fbar|)$. Feasibility uses $10^{-6}$, scaled by row for deterministic rows. Agreement on infeasibility, incomplete comparisons, and numerical disagreements are counted separately. The first D2 disagreement at the baseline is re-solved with presolve disabled, and both results are stored. This automatic recheck does not rescale constraints or tighten solver tolerances, and does not apply to the parameter trials.

\parhead{Statuses and certificates.} A check ends as \emph{flag}, \emph{no flag}, \emph{incomplete}, or \emph{not available}. A report records the version of the model and the program, the parameter values, each decision examined, the solver status, the two values, and the available witness or bounds. A numerical disagreement is a verified counterexample only when the witness or separated bounds of Proposition~\ref{prop:counter} hold at the same decision. Assessing a flag therefore requires checking solver tolerances, scaling, and the decision actually used on each side. One verified counterexample suffices even if other comparisons are incomplete.

\subsection{Level M}\label{ec:levelM}

Level M checks that each set is nonempty when the corresponding feasibility problem is available. A decision with a finite extended cost witnesses feasibility. For eligible affine rows, a phase-one LP minimizes a common violation over the decision bounds, using the vertices of the uncertainty set when they are available, and otherwise drops the uncertain rows and uses the deterministic relaxation. Infeasibility, or a strictly positive minimal violation, of a valid relaxation certifies that the original problem is infeasible. A candidate with zero violation in a relaxation still requires checking the uncertain rows and integrality. Without a certificate or a feasible witness, level M reports that feasibility has not been established.

\subsection{Walkthrough Program}\label{ec:listing}

Listing~\ref{lst:alloc} shows the model-building part of the allocation program of Section~\ref{sec:walkthrough}. Fixing $x$ leaves the free vector $w$ and the nonnegative scalar $t$ to be optimized. The coefficient of $x_j$ in row $i$ is $L_{ji}$ for the lower Cholesky factor, which implements $w=L^\top x$. The program used to test a transpose error uses $L_{ij}$ instead.

\begin{lstlisting}[style=paperpython, float=htbp, caption={Allocation program in Gurobi (model-building excerpt).}, label={lst:alloc}]
L = np.linalg.cholesky(Q)                          # Q = L L^T
x = m.addVars(n, lb=0.0, ub=1.0, name="x")         # original decisions
w = m.addVars(n, lb=-GRB.INFINITY, name="w")       # free linking variables
t = m.addVar(lb=0.0, name="t")                     # deviation, t >= 0
m.addConstr(gp.quicksum(x[j] for j in range(n)) <= 1.0)
for i in range(n):
    loading = gp.quicksum(L[j, i] * x[j] for j in range(n))
    m.addConstr(w[i] == loading)                   # w = L^T x
norm_squared = gp.quicksum(w[i] * w[i] for i in range(n))
m.addQConstr(norm_squared <= t * t)                # ||w||_2 <= t
nominal_return = gp.quicksum(c[j] * x[j] for j in range(n))
m.setObjective(nominal_return - t, GRB.MAXIMIZE)
\end{lstlisting}
\FloatBarrier

\subsection{Numerical Flags}\label{ec:threshold-diagnostic}

\parhead{A flag that disappears on rescaling.} For the allocation program at shape matrix $0.01Q$ and decision $(0.01204039664,0.22437217141)$, the norm penalty is $0.00090421060$ and the original value is $-0.01824960278$, but the fixed program returned $-0.01915381338$ with $t=0$. Its squared-cone residual, $8.18\times10^{-7}$, lies below the feasibility tolerance of $10^{-6}$, although the objective differs by $9.04\times10^{-4}$. Scaling the quadratic row by $10^6$ reduced the difference to $2.69\times10^{-9}$, and tightening the solver tolerances with the original scaling reduced it to $2.25\times10^{-11}$. The closed-form norm identifies the discrepancy as numerical. This was the only flag among the correct controls in Section~\ref{ec:check-study}.

\parhead{Decisions at a bound.} A requested decision within $10^{-6}$ outside a variable bound can be moved to the bound when the program is fixed, while the original model is evaluated at the requested point. The workflow experiments used this convention. With a steep cost function, the two evaluations can differ even for identical models. In a separate test with identical models, evaluating both costs at the bound removed the discrepancy. This test establishes agreement at the bound only. A counterexample under Proposition~\ref{prop:counter} requires evidence at a common decision. A decision outside a bound by more than the tolerance remains infeasible in the program.

\subsection{Re-Solving after a Change}\label{ec:after-protocols}

\parhead{Scenarios.} A request at the Results step creates a scenario that records the baseline, each changed datum with the code input it is linked to, and the identities and values of fixed decisions. A datum can be changed only if it is linked to a unique input of the code. Fixing a decision requires locating its solver variable and choosing a value within its bounds. The retained program is executed with the new data and with the bounds of fixed variables set to their values. The original solution remains available if the user discards the scenario.

\parhead{Premises.} Before re-solving, the system evaluates the premises that a built-in catalog records for the original model's family, for example nonnegative interval half-widths and budget for budgeted RO, a symmetric positive definite shape matrix and nonnegative radius for ellipsoidal RO, and a nonnegative radius and samples inside the support for Wasserstein DRO. A premise that fails, or whose truth cannot be determined from the data, sends the request back for review. Parameters that set the level of protection, such as a budget, a radius, or a variance bound, require the user's confirmation. Conditions that a derivation relies on beyond the catalog, such as nonnegative decisions for an endpoint rule, appear in the selected rules. The modeler reviews these additional conditions before relying on the revised solution.

%% file: appendices/ec_memory.tex
\section{Memory (Supporting Section~\ref{sec:memory})}\label{ec:memory}

\subsection{How the System Selects Entries}\label{ec:schema}

Table~\ref{tab:schema} summarizes the information a modeler needs to judge where a rule applies and where it came from. In the experimental entries, conditions and exclusions are written within the rule text rather than stored as separate fields. For an original model, the system retrieves Math entries for the problem family and Code entries for the family and solver. Within that selection, the LLM identifies the rules whose stated conditions fit the model. The model record identifies the library and the entries supplied, so the modeler can trace a derivation even after editing a personal library.

\begin{table}[h]
\centering\small
\caption{Information in a memory entry.}
\label{tab:schema}
\begin{tabularx}{\linewidth}{@{}P{3.2cm}L@{}}
\toprule
Information & Content \\
\midrule
Identity and revisions & entry identity and revision history \\
Type & Math for reformulation or Code for implementation \\
Scope & applicable families and subtypes, such as a moment condition or ground norm, and the relevant solver for Code entries. An unrestricted entry applies to all tasks. \\
Rule & mathematical transformation or implementation instruction \\
Conditions & assumptions about the model and data under which the rule applies \\
Exclusions & cases in which the rule must not be applied \\
Source & textbook, published rule, or modeler's instruction. A Code entry can also identify the Math rule it implements. \\

\bottomrule
\end{tabularx}
\end{table}

\subsection{Memory Content and Implementation}\label{ec:mem-protocol}

\parhead{System-provided memory.} We wrote one Math and one Code entry for each of the four families studied, drawing on textbooks and research papers. The polyhedral RO entry gives the LP dual for an uncertain row and treats the absolute values that arise in budgeted uncertainty with signed decisions. The ellipsoidal RO entry distinguishes a factor matrix from a shape or precision matrix and gives the support function in each case. The two DRO entries describe how moment conditions and restrictions on the support enter a worst-case expectation, and how the ground norm and support multipliers enter a Wasserstein dual. The accompanying Code entries cover Gurobi conventions for free variables, norm epigraphs, and second-order cones. They also tell the modeler which terms to compare after a discrepancy.

\parhead{Expert rules.} For polyhedral and ellipsoidal RO, the comparison library draws on Prompts 2.1, 2.3, and 2.4 of \citet{bertsimas2025robust}. For moment and Wasserstein DRO, we supplied textbook-based rules. Section~\ref{sec:eval-memory} distinguishes results for the published RO guidance from those for our DRO additions.

\parhead{Checking conditions before a re-solve.} The eight experimental entries express their assumptions in the text of the rule. They contain no executable entry-specific tests or separate lists of exclusions. When a parameter changes, the system checks the family conditions described in Section~\ref{ec:after-protocols}. A failed condition or an affected condition that cannot be evaluated sends the revision back for review. The modeler checks any additional assumptions required by the selected rule.

\subsection{A Defective Entry and Its Propagation}\label{ec:defective}

\parhead{The omitted bound on a dual variable.} The rule concerns type-1 Wasserstein DRO with the $\ell_1$ metric, so that the dual norm is $\ell_\infty$, and support $\xi\ge 0$, so that $C=-I$ and $d=0$ in Section~\ref{ec:derivations}. The dual constraints of sample $i$ and piece $k$ then read $s_i\ge b_k+a_k^\top\hat\xi_i+\gamma_{ik}^\top\hat\xi_i$ and $\|\gamma_{ik}+a_k\|_\infty\le\lambda$ with $\gamma_{ik}\ge 0$, where $s_i$ is the epigraph variable of the sample. Writing $\beta_{ik}=a_k+\gamma_{ik}$, they become
\[
s_i\ge b_k+\beta_{ik}^\top\hat\xi_i,\qquad \beta_{ik}\ge a_k,\qquad -\lambda\mathbf 1\le\beta_{ik}\le\lambda\mathbf 1 .
\]
The entry kept $\beta_{ik}\ge a_k$ and $\beta_{ik}\ge-\lambda\mathbf 1$ but omitted $\beta_{ik}\le\lambda\mathbf 1$. With the bound, the conditions on $\beta_{ik}$ are feasible only if $\lambda\ge\max_j(a_{kj})^+$, and since $\hat\xi_i\ge0$ the epigraph row is tightest at $\beta_{ik}=\max\{a_k,-\lambda\mathbf 1\}$ componentwise. Without it, every $\lambda\ge0$ is feasible, so $\lambda$ can be lowered and the program can understate the worst case. When every slope $a_{kj}(x)$ is nonpositive, as for a loss that does not increase in $\xi$, the correct condition is also $\lambda\ge 0$, and the omission is invisible.

\parhead{Effect on the solution.} In an earlier Wasserstein example, the system selected a rule that omitted the upper bound on $\beta_{ik}$. The generated equations and solver program both repeated the omission. The program returned $-2.905964$. Adding only the 15 missing bounds raised the optimum to $-2.076412$. Solving both programs again reproduced the two values without calling the LLM. The defective rule belonged to a different library from the one evaluated in Section~\ref{sec:eval-memory}.

\subsection{Applicability of the Experimental Rules}\label{ec:default-text}

The results in Section~\ref{sec:eval-memory} use the eight entries as originally written. We subsequently clarified their applicability conditions to help users decide when to use or adapt each rule. These clarifications are described below; they do not change the entries used in the reported experiments.

\parhead{Polyhedral RO.} The compact single-budget formula applies to the stated centered perturbations with coordinate bounds. Shifted sets, general absolute-value constraints, or multiple budgets fall outside that formula and require the corresponding support-function dual. For bounds $l\le\xi\le u$, this dual retains the endpoint coefficients $u_j$ and $-l_j$, even when an interval does not contain zero and one of these coefficients is negative. Deterministic equalities pass through unchanged. An equality required to hold for every uncertain realization instead requires two opposing robust inequalities.

\parhead{Ellipsoidal RO.} The factor-image representation allows a rectangular or rank-deficient matrix, whereas the inverse-quadratic representation requires a positive definite matrix. For a singular shape matrix, the factor-image form avoids an invalid matrix inverse. In either representation, the direction of an explicit minimum or maximum must be read together with the inequality direction.

\parhead{Moment DRO.} The full-space mean--variance rule applies directly with strictly positive variance bounds. For the stated positive-radius ball and ellipsoid templates, an interior mean gives an attained dual. Zero variance bounds or fixed means on curved support boundaries can prevent attainment in an unreduced dual. For example, on the unit disk, $\mathbb E\xi=(1,0)$ forces $\xi=(1,0)$ almost surely. The worst expectation of $|\xi_2|$ is zero, while the unreduced support-function dual approaches zero without attaining it. Before using an exact epigraph, zero-variance coordinates are substituted by their means; other boundary cases require reduction or a check of dual attainment.

\parhead{Wasserstein DRO.} The stated type-1 formula applies to a maximum of affine loss pieces and assumes every positive-mass empirical atom lies in the support. It uses the dual of the ground norm and retains all support terms. At zero radius, the worst expectation reduces to the empirical expectation and can be evaluated directly. An explicitly finite support requires a finite-scenario formulation; replacing it by its convex hull can enlarge the ambiguity set.

%% file: appendices/ec_ledger.tex
\section{Experimental Methods and Additional Results (Supporting Section~\ref{sec:study})}\label{ec:ledger}

The workflow experiments used the base LLM with reasoning enabled and No memory. An LLM reviewed the equations and could return them for revision before coding. In the memory experiments this review was advisory, so coding could proceed after an unfavorable assessment. The programs evaluated by the numerical check used the memory and solver configurations specified for each experiment. The independent numerical comparison itself used no LLM. The 64-task and 256-task sets draw on the same four model families but contain separately constructed instances.

\input{appendices/ec_task_construction}

\subsection{The Workflow Suite}\label{ec:suite}

\parhead{Tasks.} Each of the 40 synthetic tasks includes numerical data and a description of the decision and its uncertain inputs (Table~\ref{tab:ec-suite}). An independent grader computes a reference optimum and evaluates the returned decisions in the original problem. For N01--N08 an independent program also computes the reference for each revised state. These calculations require no LLM.

\begin{table}[!hbp]
\centering
\caption{The workflow suite.}
\label{tab:ec-suite}
\scriptsize
\setlength{\tabcolsep}{3pt}
\begin{tabularx}{\textwidth}{@{}P{1.1cm}LP{1.1cm}LP{1.1cm}L@{}}
\toprule
Task & Key feature & Task & Key feature & Task & Key feature \\
\midrule
\multicolumn{6}{@{}l}{\emph{Polyhedral RO}} \\
R02 & signed positions; absolute values needed & R04 & integer coverage; robust MILP & R05 & fractional budget $\Gamma=1.5$ \\
R10 & asymmetric coupled polyhedron & R21 & signed net positions, fractional budget & R22 & one shock on supply and limit sides \\
R23 & coupled shocks on three rows & R24 & ten suppliers, 20 decisions, coupled budget & N01 & open choice: box or budget \\
N05 & 50 suppliers; post-solve changes & & & & \\
\multicolumn{6}{@{}l}{\emph{Ellipsoidal RO}} \\
E01 & spherical ellipsoid; interior solution & E02 & factor $L$, shape $Q$, precision $M$ & E03 & precision-matrix input \\
E04 & shared shock in coefficients and right side & E21 & matrix read as shape & E22 & same matrix read as precision \\
E23 & signed rows, off-diagonal ellipsoid & E24 & 24 resources, correlated costs & N02 & open choice: joint factor radius \\
N06 & 50 generators; post-solve changes & & & & \\
\multicolumn{6}{@{}l}{\emph{Moment DRO}} \\
M04 & shifted ellipsoidal shock & M10 & mean and absolute deviation & M11 & joint capacity, coordinatewise deviation \\
M13 & mean and variance on full space & M21 & exact mean & M22 & mean upper bound \\
M23 & one joint distribution for two areas & M24 & 20 items, coupled capacity, deviation & N03 & open choice: exact mean, variance bound \\
N07 & 50 warehouses; post-solve changes & & & & \\
\multicolumn{6}{@{}l}{\emph{Wasserstein DRO}} \\
W02 & full-space portfolio, $\ell_2$ metric & W05 & hinge loss, nonnegative support & W07 & box support, large radius \\
W10 & triangular support; mass splitting & W21 & finite candidate support & W22 & convex hull of the same points \\
W23 & nonuniform reference, Euclidean transport & W24 & 24 regions, 32 samples, mass splitting & N04 & open choice: scenario support \\
N08 & 80 lanes; post-solve changes & & & & \\
\bottomrule
\end{tabularx}
\end{table}

\parhead{Protocol.} We ran the workflow with LLM at high reasoning, Gurobi, full review, and No memory. Within an attempt the system allows one retry of the analysis, one retry of a rejected reformulation, two repairs of code that fails to run, and two repairs following a counterexample. A repair supplies the previous output and the diagnostic feedback to the LLM for revision. The system then checks the revised program. A task stops as soon as a solve completes with an unflagged, complete check or uses its maximum of three attempts. The grader reads the reference only after the run. All 40 cases passed on their first attempt without code or check repairs, and the independent grader verified feasibility, cost, and optimality. We also replayed the interactions and recovered the same displayed results in all 40 cases.

\parhead{Open protection choices.} For N01--N04 the experimental inputs comprise an open description followed by a clarification of the protection requirement. N01 specified hard individual bounds with a budget of 1.5, and N02 specified a factor ball of radius 1.6. N03 fixed the means, bounded the variances, and selected the listed finite support. N04 selected twelve planning scenarios as support, ten equally weighted observations, and a mean transport distance of at most 3. All four prompted clarification. N01, N02, and N04 yielded correct models and solutions. In N03 the LLM review identified reversed holding and shortage costs in the original objective and stopped the run. The subsequent user-edit continuation in Table~\ref{tab:edit-cases} corrected that objective and yielded an independently verified solution. Thus the initial open-choice runs produced three solutions; completion of N03 required the later edit. In the separate 40-task batch, the descriptions included these choices from the start.

\parhead{Input and solver controls.} The input controls retain stated or recommended modeling choices and supply no additional values when numerical data are missing. Five underspecified inputs prompted clarification. Five invalid inputs prompted correction, including a Wasserstein radius of 0.5 when moving the sample at 2 to support $[0,1]$ requires distance at least 1. The remaining two controls returned infeasible and unbounded solver statuses, respectively.

\parhead{Completion and cost.} All ten completed-input cases in each family met $\OA$ and $\EP$ on their first attempt. The 40 cases required 283 model responses at an estimated token cost of \$2.51. Mean elapsed time was 8.46 minutes per task and median time was 8.25 minutes.

\parhead{N02 under two protection requirements.} The chosen factor ball gives a protection term of $1.6\|L^\top x\|_2$ and an optimal worst-case cost of CNY~8,389.46. An offline comparison used independent factor bounds, changing the protection term to $1.6\|L^\top x\|_1$. At the allocation chosen for the ball, worst-case cost under the box is CNY~10,335.67. The box-optimal allocation moves 7.77 of the 20 blocks, fills sites C and D, and costs CNY~9,760.62.

\parhead{Cumulative changes.} Each of the 16 requests restated all cumulative changes against the retained baseline. Fixed quantities for groups A and B (ten decisions each) came from the predeclared original case, independent of the new run. Protection changes included $\Gamma$ from 6.5 to 15.5 in N05, $\rho$ from 1.5 to 4 in N06, and $\varepsilon$ from 0.22 to 0.85 in N08. All four baselines and 16 revised states met their own independent feasibility, objective, and fixed-quantity checks. N05 and N06 reused code at all four states. N07 regenerated all four because the shortage penalties $b_i$ were not exposed as adjustable inputs in the code. The last three states also changed $V$, the variance bound on the operating stress index, which likewise lacked an adjustable input. N08 reused the first state and regenerated the final three because the Wasserstein radius $\varepsilon$ was not exposed as an input. These seven requests returned to the original-model stage and generated new reformulations and programs. They therefore did not test the premises for reusing the baseline program. We evaluated all revised solutions independently. No further pointwise check was requested.

\parhead{Additional base LLM.} GPT-6.1 Sol used medium reasoning and No memory with the same 40 inputs, clarification choices, solver, and stopping rule (Table~\ref{tab:workflow-models}). N03 initially produced a reformulation that the coding stage could not read and succeeded on its second attempt. The other 39 tasks succeeded on their first. Three outputs needed format corrections; none required repair after an execution error or a pointwise check.

Some GPT runs resumed after corrections to the workflow implementation. Recorded responses were replayed only when the complete API request matched, including the model, reasoning settings, and prompt. The evaluation used 247 replayed responses and 40 newly generated responses, all from GPT-6.1 Sol with medium reasoning. A separate replay of the 283 DeepSeek responses reproduced the DeepSeek outcomes on the corrected implementation. The 40 new GPT responses covered all seven calls for each of N02 and N05--N08, plus five calls for W10. Across 41 task attempts, the 287 distinct GPT responses comprise 280 calls in successful paths, two format corrections in M24 and W05, and five calls in N03's unsuccessful first attempt.

Table~\ref{tab:workflow-models} counts each response once at its original token usage and tariff. The GPT estimate of \$17.44 includes \$13.33 for retained responses and \$4.11 for new ones. Local replay incurs no additional API charge and differs from the provider's input-token caching, whose charges follow the original usage records. The DeepSeek estimate is \$2.51. Mean API time sums original call latencies over all evaluated attempts and divides by 40, giving 8.43 minutes for DeepSeek and 3.55 for GPT; local replay time is excluded. The totals exclude 23 additional GPT responses used for implementation diagnostics rather than the evaluated task paths.
\begin{table}[htbp]
\centering\small
\caption{Workflow configurations on the same 40 completed-input cases, both using No memory.}
\label{tab:workflow-models}
\begin{tabularx}{\linewidth}{@{}Lcrrrrr@{}}
\toprule
Model & Reasoning & First-attempt $\EP$ & Final $\EP$ & Responses & Cost (USD) & API min./task \\
\midrule
DeepSeek-V4.1-Flash-1017 & high & 40 & 40 & 283 & 2.51 & 8.43 \\
GPT-6.1 Sol & medium & 39 & 40 & 287 & 17.44 & 3.55 \\
\bottomrule
\end{tabularx}
\end{table}

\parhead{Editing and customization.}\label{ec:edit-cases} Table~\ref{tab:edit-cases} compares three cases resumed from the point of a user edit, with time measured from that point. In N03 the modeler corrected the displayed objective from $h_i(\xi_i-x_i)^++s_i(x_i-\xi_i)^+$ to $h_i(x_i-\xi_i)^++s_i(\xi_i-x_i)^+$. With reasoning enabled and No memory, the corrected model returned $(6,4,9,7)$ with worst-case cost 42.125. An independent LP over 81 joint outcomes confirmed this value. In the polyhedral case, the modeler edited the reformulation and obtained an optimum of $-7.63413402994$. The revised program passed the external evaluation, including the comparison at the reference decision. In the third case, the analyst added a Code entry to a personal library and retained the system-provided Math entries, with reasoning disabled. The generated code followed the requested builder and model names but omitted the requested prefix in its printed output. One correction to the structure of the generated response preceded the final check.

\begin{table}[htbp]
\centering\small
\caption{Solutions after editing the model, reformulation, or Code memory.}
\label{tab:edit-cases}
\begin{tabularx}{\linewidth}{@{}Lrrrr@{}}
\toprule
Continuation & Objective & Completed D2 points & LLM calls & Seconds \\
\midrule
Personal Code convention & 5.414214 & 205 & 6 & 41.35 \\
N03 original-model edit & 42.125000 & 221 & 5 & 472.93 \\
Polyhedral counterpart edit & $-7.634134$ & 205 & 3 & 43.58 \\
\bottomrule
\end{tabularx}
\end{table}

\subsection{The Memory Benchmark}\label{ec:membench}

\parhead{Reference calculations.} The 64-model collection described in Appendix~\ref{ec:task-construction} uses references computed with RSOME and a cutting-plane method driven by the check's worst-case routines, which account for mass sent to infinity. All 64 instances satisfied the reference tests. The largest cutting-plane gap relative to $1+|v_{\mathrm{LB}}|$ was $9.83\times10^{-7}$, below $10^{-6}$, where $v_{\mathrm{LB}}$ is the cutting-plane lower bound. The method took 3 to 91 iterations (median 9).

\parhead{Configurations on the 64-task set.} Table~\ref{tab:memories} compares results on the same individual tasks. With reasoning disabled, 42 tasks succeeded only with system-provided memory in its comparison with No memory, and two succeeded only with No memory. For system-provided memory versus expert rules, the corresponding counts were 23 and two. Relative to No memory, expert rules gave 26 gains and seven losses. The breakdown was three gains and seven losses for RO, and 23 gains and no losses for DRO. The comparison library uses published guidance for the two RO families and our extensions for DRO. With reasoning enabled, the total rose from 62 to 63 of 64 when system-provided memory was added. The remaining program was mathematically correct but triggered the numerical flag described below.

\begin{table}[t]
\centering\small
\caption{Configurations on 64 original models, one attempt each. Family columns count evaluation passes out of 16.}
\label{tab:memories}
\resizebox{\linewidth}{!}{\begin{tabular}{@{}llrrrrrrrrrr@{}}
\toprule
Memory & Reasoning & Polyh. & Ellips. & Moment & Wass. & $\EP$/64 & $\OA$/64 & First flagged & Calls (median) & Cost (USD) & Min./task \\
\midrule
No memory & off & 2 & 13 & 0 & 0 & 15 & 15 & 45 & 9 & 1.74 & 2.10 \\
Expert rules & off & 1 & 10 & 12 & 11 & 34 & 35 & 32 & 8 & 1.50 & 1.72 \\
System-provided & off & 10 & 15 & 15 & 15 & 55 & 55 & 11 & 4 & 1.03 & 1.20 \\
No memory & on & 16 & 15 & 16 & 15 & 62 & 62 & 3 & 4 & 3.04 & 5.74 \\
System-provided & on & 16 & 16 & 16 & 15 & 63 & 64 & 2 & 4 & 2.54 & 5.20 \\
\bottomrule
\end{tabular}}
\end{table}

\parhead{Separate 256-task set.} Each family contributes 64 models, with two to five decisions in RO and two to eight in DRO. Table~\ref{tab:memory-final4} compares three guidance choices with reasoning disabled and a fourth configuration with reasoning enabled and No memory. All four use the base LLM, the same implementation, and the same numerical inputs. Two independent formulations, one explicit and one native to the modeling package, were used to check the 256 reference values within $10^{-5}(1+|f^\star|)$. No numerical instance was common to the 64-task and 256-task sets. The controlled-change study uses these same 256 base models, with one deliberate alteration to each model.

\parhead{Independent selection of Math and Code entries.} Table~\ref{tab:stage-selection} compares four choices on the same 64 tasks with reasoning disabled. The Math-only condition supplies mathematical guidance for the reformulation and any check-guided repair. The Code-only condition supplies implementation guidance when generating the solver program. Both conditions allow one attempt per task, including internal retries, across all 64 tasks. Math-only gave $\EP=50$ and $\OA=51$ in 333 calls. Code-only gave $\EP=13$ and $\OA=13$ in 553 calls. Mean elapsed time and cost include all 64 tasks. The results for neither and both types of guidance come from the earlier comparison in Table~\ref{tab:memories}. We subsequently ran the two single-stage configurations on the same tasks, using a common implementation and the same external grading criteria. Some of the mathematical and coding instructions overlap, so the comparison measures the effect of supplying each group of entries as written.

\parhead{GPT memory configurations.} Table~\ref{tab:terra-memory} compares three GPT-6.1 Terra configurations on the same 64 models. Each used advisory review, the same repair limits, four concurrent tasks, and one attempt per task. All three were evaluated at the reference decision as well as the ordinary trial decisions. System-provided memory was used with reasoning disabled, and No memory was used with medium or high reasoning. The experiment therefore compares complete configurations. In all three, $\OA=\EP$. With memory, all 48 RO and moment DRO tasks and 13 of the 16 Wasserstein tasks passed. The remaining three stopped before producing a usable input to the coding stage. With reasoning enabled and No memory, all 64 passed. None of the three configurations triggered a check-guided repair.

\begin{table}[htbp]
\centering\small
\caption{GPT-6.1 Terra configurations on 64 models, one attempt per task including permitted repairs.}
\label{tab:terra-memory}
\begin{tabularx}{\linewidth}{@{}Lcrrrrr@{}}
\toprule
Guidance & Reasoning & $\EP$/64 & Calls & Cost (USD) & Case min./task & API min./task \\
\midrule
System-provided & none & 61 & 261 & 7.45 & 1.58 & 1.40 \\
No memory & medium & 64 & 256 & 8.09 & 2.58 & 1.92 \\
No memory & high & 64 & 257 & 9.54 & 3.04 & 2.84 \\
\bottomrule
\end{tabularx}
\end{table}

Case time includes recorded network waits and local computation. API time sums the original latencies of returned model calls. Queue pauses and offline grading are excluded. Waiting for failed requests added 571.87 seconds in the high-reasoning group, 2,466.16 in the medium group, and 541.71 in the system-provided group, summed over all 64 cases in each. For Terra, the rates per million tokens were \$1 for ordinary input, \$0.05 for cache hits, \$1.25 for cache writes, and \$5 for output. Sol uses twice those rates. Cache-write tokens are excluded from ordinary input, and reasoning tokens are already included in output. The Sol estimates price only the token usage returned by the provider.

\parhead{External grading and numerical flags.} The evaluator constructs the original model from the benchmark arrays, runs S, D1, and D2 with $k=200$ and seed 1, and compares projected values at the reference optimizer. The reference optimizer is available only during external grading. A program passes evaluation when it matches the optimal value and the audit completes without a flag at any tested decision.

Two programs on the 64-task set illustrate different reasons for an audit flag. With expert rules and reasoning disabled, a Wasserstein program matched the optimal value but omitted the box support. At an interior decision it gave 1.1827229, exceeding the independently verified original dual upper bound of 1.1793091. With system-provided memory and reasoning enabled, the only value-matching program that did not pass evaluation had a gap of $1.84\times10^{-5}$ at one probe, just above tolerance $1.74\times10^{-5}$. Turning off presolve restored agreement. The stated scoring rule counts the initial numerical flag and gives $\EP=63$.

\parhead{Runtime checks and external evaluation.}\label{ec:runtime-evaluation} Table~\ref{tab:runtime-outcomes} partitions the 256 final outcomes for each configuration. We classified a delivery as clear when the solve completed and all requested D2 comparisons finished without a flag. Four clear deliveries failed external grading, three with No memory and reasoning disabled and one with system-provided memory. All four passed their ordinary D2 probes, missed the reference objective tolerance, and disagreed at the reference optimizer. In the system-provided case, a polyhedral RO program reported 20.969268 against 27.711220, a 24.3\% difference between reported and reference objectives. This discrepancy measures the error in the reported objective; we did not measure the economic loss from implementing the returned decision. Flags in the other columns include numerical discrepancies that have not been individually adjudicated.

\begin{table}[htbp]
\centering\small
\caption{Final outcomes of the 256-task memory runs. The five numeric columns partition each configuration.}
\label{tab:runtime-outcomes}
\begin{tabularx}{\linewidth}{@{}Lcrrrrr@{}}
\toprule
 & & \multicolumn{2}{c}{Completed, runtime clear} & \multicolumn{2}{c}{Completed, runtime not clear} & \\
\cmidrule(lr){3-4}\cmidrule(lr){5-6}
Memory & Reasoning & $\EP$ & Not $\EP$ & Flagged & Incomplete & No completed solve \\
\midrule
No memory & off & 81 & 3 & 43 & 1 & 128 \\
Expert rules & off & 188 & 0 & 19 & 2 & 47 \\
System-provided & off & 241 & 1 & 4 & 0 & 10 \\
No memory & on & 255 & 0 & 0 & 0 & 1 \\
\bottomrule
\end{tabularx}
\end{table}

The first check was flagged on 148, 88, 21, and zero tasks in these four configurations. After the permitted repairs, 27, 41, nine, and zero of those tasks passed evaluation, respectively. These counts follow each flagged program through repair within the same attempt. Final success totals also include programs whose first checks were incomplete or unavailable. The separate retry comparison below uses eleven cases from the 64-task set.

\parhead{Actual check coverage.} The four 256-task configurations ran the check 409, 379, 282, and 258 times, including rechecks within an attempt. All used the baseline data and requested M, S, D1, and D2. Across these calls, the numbers of completed D2 comparisons were 78,580, 72,983, 57,631, and 53,831, with 7,144, 6,778, 1,856, and 620 requested comparisons reported incomplete. Repeated checks of the same task and decision are counted each time. Restricting to final clear deliveries gives 17,832, 39,580, 51,070, and 53,831 complete comparisons. The external memory evaluation also uses the baseline data and adds the reference decision. Section~\ref{ec:check-study} evaluates \Dt{} separately on programs with adjustable parameters.

\parhead{Time and token estimates.} Times use elapsed duration of the recorded first attempt for every task, including unsuccessful tasks and automatic repairs. Queue pauses, separate preliminary runs, and offline grading are excluded. In the 256-task runs, total calls are 2,068 without memory or reasoning, 1,596 with expert rules, 1,151 with system-provided memory, and 1,042 with reasoning alone. Total elapsed task times are respectively 489.34, 353.49, 177.83, and 1,452.46 minutes. Cost differences include the number and length of repair calls and reasoning output, not simply the added entry text.


\subsection{The Check Study}\label{ec:check-study}

\parhead{Archived programs.} The audit covered 68 workflow and 32 one-request programs. Of 100 programs, 86 solved to optimality and 74 matched the reference objective. The benchmark grader compared values at the reference decision, the returned decision when available, their midpoint, and forty further decisions (usually 43 points, once 42). Among the 74 objective matches, 63 completed without a flag, ten were incomplete, and one was refuted. That moment DRO program reported the optimum 2 at a decision whose original worst-case cost was $8/3$.

\parhead{Prior assessments.} Before developing the check, the authors assessed the reformulation and code for 32 workflow and 32 one-request programs using the task specifications and references. Among eight assessed as defective, D2 flagged five and was incomplete on three. D0 and D1 each flagged four, together covering the same five. Among the 56 assessed as correct, D2 completed without a flag on 48 and was incomplete on eight.

\parhead{Controlled changes.} The modifications in Table~\ref{tab:mutations} represent omissions or misinterpretations of uncertainty information. They make robust constraints nominal, transpose ellipsoid factors, relax mean or variance conditions, widen supports, or change transport norms. Each design tests the changed model's optimizer, four Gaussian neighbors, the available box corners, and $k=40$ further decisions from the box and the nominal relaxation. All 256 unchanged copies agreed. The comparison below covers 211 non-equivalent changes with finite optima. It excludes 41 changes classified as equivalent in the controlled study and four changed models without finite optima.

With trial decisions centered on the changed optimizer, D2 detected 181 changes. Centering on the original optimizer gave 175 detections, and combining both sets gave 192. Under the changed-optimizer protocol, 75 changes preserved the optimal value, of which D2 detected 56. Restricting to the 74 changes missed by both D0 and D1 left 55 D2 detections. Combining D0 with D2 detected 192 of 211 changes, with row totals of 45, 33, 39, 19, 2, 36, and 18 in Table~\ref{tab:mutations}. This count comprised 102 changes flagged by all three checks, 23 by D0 and D2 only, eleven by D0 only, one by D1 and D2 only, and 55 by D2 only. Nineteen escaped all three checks.

The offline boundary experiment used a policy that bisects the segment from the most recent commonly accepted decision toward a decision rejected by both models, up to twelve times. Over five seeds, it detected 40--46 of the 49 ellipsoidal changes at a budget of 50 decisions, compared with 19--21 for mixed sampling. At 200 decisions it detected 48--49. On the 48 polyhedral changes it detected 40--44, compared with 45--47 for mixed sampling. The workflow runs use the sampling procedure in Section~\ref{ec:checker}.

\parhead{Executable controls.} Twelve deliberately flawed and seven correct programs were checked at 200 random decisions plus structured points. Where adjustable parameters existed, each program was also tested at 20 random parameter vectors and at specified boundary or structural settings. On the flawed programs, baseline D2 flagged eight. Parameter changes exposed the remaining four. One program reduced an interval support to three points, two removed a sign restriction on a deviation or budget multiplier, and one omitted the nonnegative support. Among the programs flagged at baseline, one matched the intended optimum of 2.5, but its decision $(2,1)$ violated the displayed model by 1 at $z=-1$. All seven correct programs passed baseline checks. Six had adjustable parameters and passed all 120 random vectors. The only numerical flag among 31 specified settings arose in the ellipsoidal allocation example at $0.01Q$ (Section~\ref{ec:threshold-diagnostic}). Each program took 0.01--6.85 seconds to check.

\parhead{A sampling miss.} Across the five configurations in Table~\ref{tab:memories}, one delivered program passed all 205 D2 probes yet failed the external reference criteria. This polyhedral program with system-provided memory and reasoning disabled rejected the reference optimizer and reported $-7.6310$ against $-7.6341$. D0 flagged the objective discrepancy. None of the 205 decisions lay in the region that separated the models.

\parhead{Repair after a flagged comparison.} Table~\ref{tab:feedback} follows the 93 task--configuration runs of Table~\ref{tab:memories} whose first checked program was flagged, spanning 58 distinct tasks. A program that returned a solution was checked at 205 decisions (the returned one, four neighbors, and 200 random ones). When a program returned no solution, it was checked at the 200 random decisions. In the repair case of Section~\ref{sec:eval-check}, D2 flagged 121 of 205 decisions in the first program, the first repair was unbounded at level S and flagged 160 of 200 decisions, and the second repair agreed at all 205. The complete attempt used 15 LLM calls and 7.7 minutes.

\begin{table}[htbp]
\centering\small
\caption{Flagged task--configuration runs and final outcomes within the same attempt.}
\label{tab:feedback}
\begin{tabular}{@{}llrrrrr@{}}
\toprule
Memory & Reasoning & Flagged & Returned a solution & $\EP$ after repair & Still flagged & Other \\
\midrule
No memory & off & 45 & 38 & 7 & 9 & 29 \\
Expert rules & off & 32 & 19 & 10 & 9 & 13 \\
System-provided & off & 11 & 6 & 5 & 4 & 2 \\
No memory & on & 3 & 2 & 3 & 0 & 0 \\
System-provided & on & 2 & 2 & 1 & 1 & 0 \\
\midrule
All & & 93 & 67 & 26 & 23 & 44 \\
\bottomrule
\end{tabular}

\smallskip
\parbox{\linewidth}{\footnotesize \emph{Returned a solution}: the first flagged program solved and returned a decision. \emph{Other}: the attempt ended without a solved program or with an incomplete check.}
\end{table}

\parhead{Retries with and without diagnostics.} We compared one continuation of each kind for the eleven tasks whose first program was flagged with system-provided memory and reasoning disabled (Table~\ref{tab:memories}). Both began at the same failure state, with the same previous counterpart, task, memory, numerical stopping checks, and limit of two repair cycles. The comparison removed the D1/D2 diagnostics and previous diagnostic review from the repair prompts. Five tasks passed evaluation with detailed feedback and four without it. Three succeeded in both conditions, two only with detailed feedback, one only without, and five in neither. Replaying the original continuations reproduced all eleven outcomes, and inspection of the new requests confirmed removal of the diagnostics. The comparison covers eleven of the 93 flagged runs, with one continuation per task and condition.

\subsection{Published Programs for a Deterministic Benchmark}\label{ec:published}

Of the 100 IndustryOR programs released for ORLM-LLaMA-3-8B, 37 reproduced the objective label within relative tolerance $10^{-4}$, allowing for rounded labels \citep{huang2025orlm}. Against our readings of the problem texts, 22 represented the intended problem and three represented a different one, including a disputed storage-cost interpretation. Of the remaining twelve, ten labels disagreed with our reference, one was disputed, and one problem lacked data \citep{zhang2025optimind}. We selected four discrepant programs and three controls to examine the distinction between checking the generated program against the displayed model $\Pdisp$ and checking it against the intended reference $\Pref$.

At the returned decision, four neighbors, and 200 random decisions, all three controls agreed with $\Pref$. Programs 7, 69, 73, and 11 disagreed with $\Pref$ on 37, 69, 38, and 19 of 205 decisions, respectively. For example, problem 7 accepted $(7,1,6)$ despite a limit of two transport modes. The three separately evaluable displayed models shared their code's omission and raised no flags at these points. Under uniform perturbations of up to $\pm50\%$ in each numerical parameter, the program and reference optima differed on 59 of 200 instances of problem 69 and five of 198 solvable instances of problem 7.

%% file: appendices/ec_task_construction.tex
\subsection{Task Construction}\label{ec:task-construction}

The workflow cases cover business descriptions, modeling choices, and post-solve changes. The two model collections vary mathematical structures and dimensions and contain distinct numerical instances.

\parhead{Workflow cases.} The suite combines 32 application cases with eight interaction cases, giving ten per family. The first group comprises sixteen existing and sixteen additional cases covering signed and integer decisions, fractional budgets, correlated perturbations, moment equalities and inequalities, and finite versus continuous support. Paired cases distinguish similar descriptions with different mathematical meanings. For example, E21 and E22 interpret the same numerical matrix as a shape matrix and a precision matrix, respectively; M21 and M22 distinguish an exact mean from a mean upper bound. Table~\ref{tab:ec-suite} lists the coverage.

Each case contains a description, data, mathematical specification, and independently checked reference. Input preparation separates the description and data from the derivation and reference answer. N01--N04 begin with open protection choices for the interaction study; their completed descriptions are used in the 40-case batch. N05--N08 provide larger procurement, generation, inventory, and transport cases with 50, 50, 50, and 80 original decisions. Changes and fixed decisions are specified before the runs. These synthetic cases are selected for coverage.

\parhead{The 256-model collection.} Each family contributes 64 instances. In both RO families, the number of decisions and the number of constraint rows each take the values two, three, four, and five, with sixteen instances at each size. Exactly one expression among the objective and constraint rows is deterministic. The remaining expressions use box, budgeted, or general polyhedral uncertainty in the polyhedral family and ellipsoidal uncertainty in the ellipsoidal family. Decision bounds may allow positive values, negative values, or both. Both minimization and maximization objectives occur.

The DRO objectives take the form
\[
\sup_{\mathbb P\in\mathcal P}\mathbb E_{\mathbb P}
\left[\max_{k=1,\ldots,K}\{(C_kx+e_k)^\top\xi+f_k^\top x+g_k\}\right].
\]
The moment models have two to eight decisions, two to five uncertain coordinates, and two to five affine pieces. Ten combinations of moment information and support occur, each six or seven times: fixed means with box, polyhedral, ball, ellipsoidal, or budgeted support; mean intervals with box, polyhedral, or ellipsoidal support; mean and absolute-deviation bounds with box support; and mean and variance bounds on full space. The Wasserstein models have two to eight decisions, two to six uncertain coordinates, one to four affine pieces, and five to 25 equally weighted samples. They cross three ground norms, $\ell_1$, $\ell_2$, and $\ell_\infty$, with four supports: full space, the nonnegative orthant, a box, and a polyhedron. Each combination occurs five or six times.

The recorded base seeds are 26102803, 26102813, 26102823, and 26102833 for polyhedral RO, ellipsoidal RO, moment DRO, and Wasserstein DRO, respectively. Reference preparation replaced three ellipsoidal and seven Wasserstein instances whose reference computations were unresolved, using new draws of the same subtype. The retained reference values agree between explicit and native formulations within $10^{-5}(1+|f^\star|)$; the largest absolute difference is $1.17\times10^{-6}$. The numerical arrays used for these calculations also supply the original equations and parameter tables shown to the LLM. Reference values and decisions are excluded from its input.

\parhead{The mixed 64-model collection.} The 64-model collection has 16 instances per family, using separately sampled parameters and larger configurations. Each RO instance has five decisions, five rows, and an uncertain objective. Each polyhedral instance contains at least two general polyhedral rows; each ellipsoidal instance contains four ellipsoidal rows and one deterministic row. Moment models have six to eight decisions, four or five uncertain coordinates, and two to five affine pieces. They combine fixed or interval means, absolute deviations, and variances with box, polyhedral, ball, ellipsoidal, budgeted, or full-space supports. Wasserstein models have six to eight decisions, four to six uncertain coordinates, two to four affine pieces, and 15 to 25 samples. The $\ell_2$, $\ell_1$, and $\ell_\infty$ groups contain six, five, and five cases, respectively, distributed over box, polyhedral, and nonnegative supports.

The generator uses NumPy's default random-number generator with base seed 26103100 and separate streams indexed by family, task, and candidate draw. For RO, it samples the decision-bound magnitude uniformly from $[1.2,9.9]$ and rounds it to one decimal place. Nominal objective coefficients are drawn on the one-decimal grid from $-9.9$ to $9.9$. Constraint coefficients are drawn on the grid from $-9$ to $9$, with additional zero coefficients. Right-hand sides are formed around a sampled decision with nonnegative slack. Budgeted perturbations have deviations from 0.1 to 1.8 and budget drawn from $[0.3,0.6d]$, where $d$ is the number of perturbed coefficients. General polyhedra combine coordinate bounds with one or two linear relations. Ellipsoid factors come from a Cholesky factorization of a rounded matrix $BB^\top+D$, where $B_{ij}\sim N(0,0.25^2)$ and the diagonal entries of $D$ are uniform on $[0.05,0.4]$.

For moment DRO, the mean coordinates are uniform on $[-1,1]$. Supports are constructed around that mean. Box half-widths range from 0.3 to 3.2, absolute-deviation bounds from 0.4 to 1.6, and variance bounds from 0.5 to 2.0. Entries of $C_k$ and $f_k$ are uniform on $[-2,2]$, and entries of $e_k$ and $g_k$ on $[-1,1]$. For Wasserstein DRO, empirical samples come from a diagonal normal distribution, with coordinate means uniform on $[-2,2]$ and standard deviations uniform on $[0.55,2]$. Absolute values give the nonnegative-support samples. Box bounds enclose the samples with a margin; polyhedral supports add two inequalities that also contain them. The radius is uniform on $[0.28,1.03]$, and all loss coefficients are uniform on $[-1,1]$. Both DRO families use $x\in[-6,6]^n$ with one or two additional linear inequalities. Their coefficients are rounded to four decimal places, and Wasserstein radii to six.

A candidate is retained only when its reference has a finite objective with magnitude at most $10^4$, the RSOME decision satisfies the original robust constraints within $10^{-6}$, and an independent worst-case evaluation reproduces its objective. A cutting-plane lower bound must also close the optimality gap within $10^{-6}(1+|f^\star|)$. The generator permits a new draw when these conditions fail. All 64 retained instances passed on the first draw. The reference computations and rendering of the numerical models use no LLM.